\documentclass[10pt,twocolumn,letterpaper]{article}

\usepackage{iccv}
\usepackage{times}
\usepackage{epsfig}
\usepackage{graphicx}
\usepackage{amsmath}
\usepackage{amssymb}
\usepackage{float}
\usepackage{multirow}
\usepackage{color, colortbl}
\usepackage{xcolor}
\usepackage{makecell}
\usepackage{etoolbox}
\usepackage[percent]{overpic}
\usepackage{xmpmulti}
\usepackage{pdfpages}
\usepackage{subcaption}

\newcommand\tstrut{\rule{0pt}{2.4ex}}
\newcommand\bstrut{\rule[-1.0ex]{0pt}{0pt}}
\definecolor{LightGrey}{rgb}{0.9,0.9,0.9}

\DeclareCaptionLabelFormat{nocaption}{}

\usepackage[pagebackref=true,breaklinks=true,colorlinks,bookmarks=false]{hyperref}
\usepackage{cleveref}

\iccvfinalcopy 

\def\iccvPaperID{***} 

\ificcvfinal\fi

\usepackage[labeled,resetlabels]{multibib}
\usepackage{xfp}
\newcommand\SupplementaryMaterials{%
  \xdef\presupfigures{\arabic{figure}}
  \xdef\presupsections{\arabic{section}}
  \renewcommand\thefigure{S\fpeval{\arabic{figure}-\presupfigures}}
  \renewcommand\thesection{S\fpeval{\arabic{section}-\presupsections}}
  \renewcommand{\thetable}{S\arabic{table}}
  \renewcommand{\theequation}{S\arabic{equation}}
}

\begin{document}

\title{Leaping Into Memories: Space-Time Deep Feature Synthesis}

\author{Alexandros Stergiou \hspace{2em} Nikos Deligiannis\\
Vrije Universiteit Brussel, Belgium \& imec, Belgium \\
{\tt\small <first>.<last>@vub.be}
}

\twocolumn[{%
\renewcommand\twocolumn[1][]{#1}%
\maketitle
\begin{center}
    \centering
    \vspace{-1.5em}
    \captionsetup{type=figure}
    \includegraphics[width=.95\textwidth, trim={.9cm .4cm .2cm .6cm},clip]{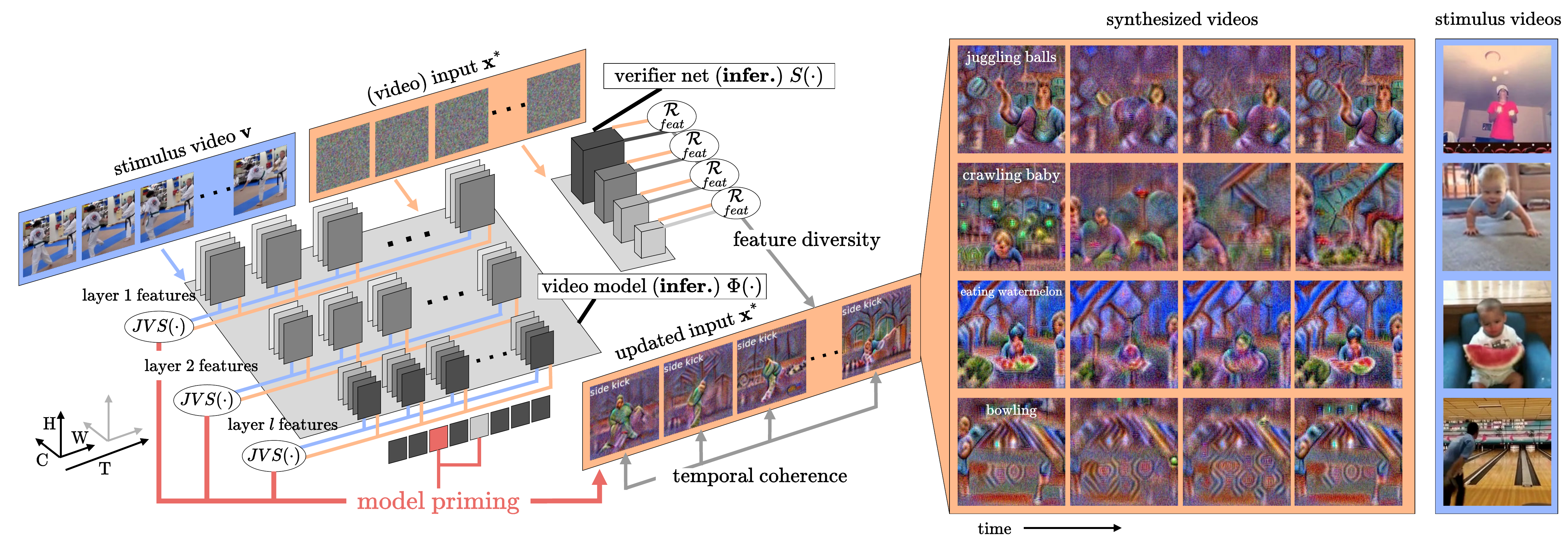}
    \captionof{figure}{\vspace{-.1em}\textbf{Illustration of our proposed LEAPS model inversion}. Given a video input $\mathbf{x}^{*}$ initialized with noise, we use a stimulus video $\mathbf{v}$ of target class $y$, to prime video model $\Phi(\cdot)$. We iteratively optimize $\mathbf{x}^{*}$ to synthesize and visually represent the internal spatiotemporal features of $\Phi(\cdot)$. To impose diversity over the synthesized videos, feature statistics from a domain-specific verifier network $S(\cdot)$ are distilled to regularize $\mathbf{x}^{*}$. To preserve the continuity of motions across frames, $\mathbf{x}^{*}$ is temporally regularized at each training iteration. Although the illustration visualizes internal representations as $C \! \times \! T \! \times \! H \! \times \! W$ volumes, they can also be flattened patches $CP^3 \times \frac{THW}{P^3}$ of $P^3$ resolution as in Vision Transformers.}
\label{fig:leaps}
\end{center}%
}]

\maketitle

\begin{abstract}
\vspace{-1.2em}
The success of deep learning models has led to their adaptation and adoption by prominent video understanding methods. The majority of these approaches encode features in a joint space-time modality for which the inner workings and learned representations are difficult to visually interpret. We propose \textbf{LEA}rned \textbf{P}reconscious \textbf{S}ynthesis (LEAPS), an architecture-independent method for synthesizing videos from the internal spatiotemporal representations of models. Using a stimulus video and a target class, we prime a fixed space-time model and iteratively optimize a video initialized with random noise. Additional regularizers are used to improve the feature diversity of the synthesized videos alongside the cross-frame temporal coherence of motions. We quantitatively and qualitatively evaluate the applicability of LEAPS by inverting a range of spatiotemporal convolutional and attention-based architectures trained on Kinetics-400, which to the best of our knowledge has not been previously accomplished. \footnote{See \href{https://alexandrosstergiou.github.io/project_pages/LEAPS/index.html}{alexandrosstergiou/LEAPS} for video examples and code.}
\end{abstract}

\section{Introduction}
\label{sec:intro}

Inverting deep networks’ learned internal representations has been a difficult task to achieve. The adaptation and deployment of CNNs and more recently Transformers, to a variety of vision tasks has led to significant breakthroughs. The field of video action recognition has experienced drastic growth in recent years through the convergence of models with increased complexities and capacities~\cite{arnab2021vivit,bertasius2021space,mangalam2022reversible,yan2022multiview} as well as large accuracy improvements~\cite{feichtenhofer2020x3d,feichtenhofer2019slowfast,li2022mvitv2,li2022uniformerv2,liu2022video}. Despite great progress in their applicability, there is still a large gap in the interpretability of video models. As these models compositionally encode space and time modalities of videos over large feature spaces and require a lot of parameters, conceptualizing their internal representations remains challenging. In this paper, we propose a method to invert learned features of video models associated with specific actions by optimizing a parameterized input to synthesize conceptually and visually coherent representations.   

In cognitive science, stages of consciousness include the conscious, the unconscious, and the preconscious. In contrast to the conscious and unconscious, the preconscious is responsible for learned information currently outside conscious awareness to remain readily available~\cite{strachey1959standard}. One way of accessing learned preconscious information and making it part of conscious awareness is through priming~\cite{Neely2003-NEEP}. Priming uses a stimulus to activate related learned concepts in memory and make them easily and readily accessible~\cite{marcel1983consciousa,marcel1983consciousb}; e.g., one can remember their bedroom if primed with a picture of a bed. 

\noindent
Motivated by visual priming in cognitive science, we demonstrate that learned representations of video models can become accessible through \emph{model priming}. By using a video stimulus and a target action class, we synthesize the dominant learned concepts corresponding to actions. In turn, the visual features of the synthesized videos provide a conceptual view of the models' learned internal representations. We term these features as the \emph{learned preconscious} of the video model associated with a specific action.   

We introduce \textbf{LEA}rned \textbf{P}reconscious \textbf{S}ynthesis (LEAPS), illustrated in \Cref{fig:leaps}, a spatiotemporal model inversion method that synthesizes interpretable videos by minimizing a classification and priming loss without prior knowledge of the training data. LEAPS uses a video of a target action class as stimulus, to prime a fixed spatiotemporal model. We additionally include two regularization terms. The first term enforces motion coherence across frames by constraining their representations at each update. The second enhances the diversity of the synthesized videos by using a domain-specific verifier network that exploits disagreements between feature statistics similar to \cite{yin2020dreaming}. Through the same architecture-independent approach, we show that LEAPS can invert the spatiotemporal features of 3D-CNNs and spatiotemporal Transformers. 

Our main contributions are as follows. 
First, we introduce LEAPS, a general approach for inverting video models, which to the best of our knowledge, is the first attempt to create videos of conceptual representations from jointly encoded space-time features.
Second, we use LEAPS on multiple convolution and attention models and compare it to prior image-based inversion methods that we extend to video.  
Finally, we show that LEAPS can invert both video CNNs and transformers, with the same architecture-independent method without any modifications.

\section{Related Work}
\label{sec:related}

Approaches for visualizing and interpreting deep models can be divided into three groups which we detail below. 

\noindent
\textbf{Attribution-based visualizations}. These methods have been used to visualize feature contributions over an input. Attribution methods for images have primarily been based on back-propagating activations of classes~\cite{chattopadhay2018grad,selvaraju2017grad,wang2020score}, or individual neurons~\cite{bach2015pixel,binder2016layer,shrikumar2017learning,springenberg2015striving} to localize regions in the input that are informative for a given class or feature. Such approaches include Integrated gradients (IG)~\cite{sundararajan2017axiomatic} which produce pixel-wise attributions from integrating computed backprop gradients. Subsequent works have also included gradient smoothing~\cite{smilkov2017smoothgrad}, gradient accumulation in saturated regions~\cite{miglani2020investigating}, and adaptation of the gradient path~\cite{kapishnikov2021guided}. Another set of approaches that rely on attribution-based visualizations use perturbations of the input~\cite{fong2017interpretable,fong2019understanding,petsiuk2018rise}. Given their straightforward applicability, image attribution methods have also been extended to video. The majority of works; e.g., Saliency Tubes~\cite{stergiou2019saliency,stergiou2019class}, STEP~\cite{li2021towards}, video OSA~\cite{uchiyama2023visually}, and BOREx~\cite{kikuchi2022borex}, have focused on the localization of spatiotemporal salient regions by extending Grad-CAM~\cite{selvaraju2017grad}, Extremal Perturbations (EP)~\cite{fong2019understanding}, Occlusion Sensitivity Analysis (OSA)~\cite{zeiler2014visualizing}, and Gaussian processes regression (GPR)~\cite{burke2017leveraging,mokuwe2020black} respectively. In contrast to these approaches that localize salient class features, we offer a visual feature synthesis method to conceptualize the learned representations of video models through priming.            

\noindent
\textbf{Input synthesis}. Network-centric approaches invert models to either visualize particular classes~\cite{nguyen2016multifaceted,yosinski2015understanding,yin2020dreaming} or features~\cite{olah2017feature,stergiou2021mind}. One of the main methods employed for visualizing internal network features is Gradient Ascent (GA)~\cite{erhan2009visualizing}, which optimizes the input by increasing the activation of a specific neuron. Activation Maximization (AM)~\cite{simonyan2013deep} later adapted GA to visualize CNN features. Following works have been built on top of AM by including additional regularizers; e.g., total variation~\cite{mahendran2016visualizing}, blurring~\cite{wang2018visualizing}, and gradient masking~\cite{nguyen2016multifaceted}. One of the most popular extensions of AM has been DeepDream~\cite{alexander2015inceptionism} which optimizes the input to yield high responses for a chosen class while keeping internal representations constraint-free. The produced images include repetitions of recognizable concepts without representing a coherent whole. In addition to AM, model inversion~\cite{dosovitskiy2016inverting,ghiasi22plug,mahendran2015understanding,yin2020dreaming,zeiler2014visualizing} includes the task of maximizing the classification score of the synthesized image instead of maximizing class activations. The only extension of input synthesis approaches to videos has been introduced by Feichtenhofer \etal~\cite{feichtenhofer2020deep} in which AM is used to create visual representations of class features from two-steam models~\cite{feichtenhofer2016convolutional,wang2016temporal}, trained on RGB frames (spatial stream) and optical flow (temporal stream). In this paper, we instead propose a model inversion method for inverting video models concurrently encoding space and time modalities.   

\noindent
\textbf{Visual feature generation}. These methods utilize activation maximization by including an additional generator network~\cite{nguyen2017plug,nguyen2016synthesizing,odena2017conditional}, with the cost of requiring access to the training data. Huang \etal~\cite{huang2018makes} adapted feature generation on video data by modeling the temporal signal with a temporal generator network. Despite the high fidelity of the produced results, these approaches are less suitable for interpreting internal model behaviors, as they do not solely depend on the model under inspection. Instead, they are primarily influenced by the training data used as well as the capacity and complexity of the generator model trained.

\section{Learned Preconscious Synthesis}
\label{sec:method}

In this section, we overview our LEAPS method, shown in~\Cref{fig:leaps}.
We start by formally introducing model priming for video models in~\Cref{sec:method::priming}. A stimulus video of a target class is used to initially prime the network. The training process uses the primed representations to update a randomly initialized input alongside two regularizers. The first regularizer is used to enforce temporal coherence, explained in~\Cref{sec:method::temp_coh}. The second is used to improve synthesized feature diversity, overviewed in~\Cref{sec:method::feat_div}. We present the final aggregated function in~\Cref{sec:method::aggregation}.


\subsection{Model Priming}
\label{sec:method::priming}

Priming deep models is influenced by cognitive science~\cite{Neely2003-NEEP} in which a stimulus is used to recall prior knowledge. We extend this approach to visualizing the learned preconscious of deep video models associated with a specific action class $y$. Given a video $\textbf{v}$ of size $C \! \times \! T \! \times \! H \! \times \! W$, with $C$ channels, $T$ frames, $H$ height, and $W$ width, as a visual cue for action class $y$, and a randomly initialized input $\mathbf{x}^{*}$ of $C \! \times \! T \! \times \! H \! \times \! W$ size to optimize. We define a priming loss $\underset{prim}{\mathcal{L}}(\mathbf{x}^{*},\mathbf{v})$ between the internal representations $\textbf{z}^{l}(\cdot)$ of the optimized input $\mathbf{x}^{*}$ and stimulus $\mathbf{v}$ across $l \! \in \mathbf{\Lambda}=\{1,...,L\}$ layers:
\begin{equation}
\label{eq:priming}
    \underset{prim}{\mathcal{L}}(\mathbf{x}^{*},\mathbf{v}) \! = \!\! \; \frac{1}{L} \sum_{l \in \mathbf{\Lambda}} \lambda_{l} \; JVS \left( \mu \!\left( \textbf{z}^{l}(\mathbf{x}^{*}) \right) \!,\mu \!\left(\textbf{z}^{l}(\mathbf{v})\right) \right) 
\end{equation}
\noindent
where $\mu \!\left( \textbf{z}^{l}(\cdot) \right)$ is the $C$-length spatiotemporal mean vector of representations $\textbf{z}^{l}(\cdot)$. $JVS(\cdot)$ denotes the Jaccard vector similarity~\cite{fernando2021anticipating}. To integrate a degree of freedom and avoid hard constraints on the internal representations of $\mathbf{x}^{*}$, we define $0 < \lambda_{l} \leq 1$ as priming weight for layer $l$. An overview of updating $\mathbf{x}^{*}$ by priming appears in \Cref{fig:leaps}. 

Due to the vastness of the feature space when optimizing (\ref{eq:priming}), we include two additional regularization terms to constrain the input. Specifically, we apply a temporal coherence regularization $\underset{coh}{\mathcal{R}}$ and a feature diversity regularization $\underset{feat}{\mathcal{R}}$, which we detail below.

\begin{figure}
    \centering
    \includegraphics[width=\linewidth]{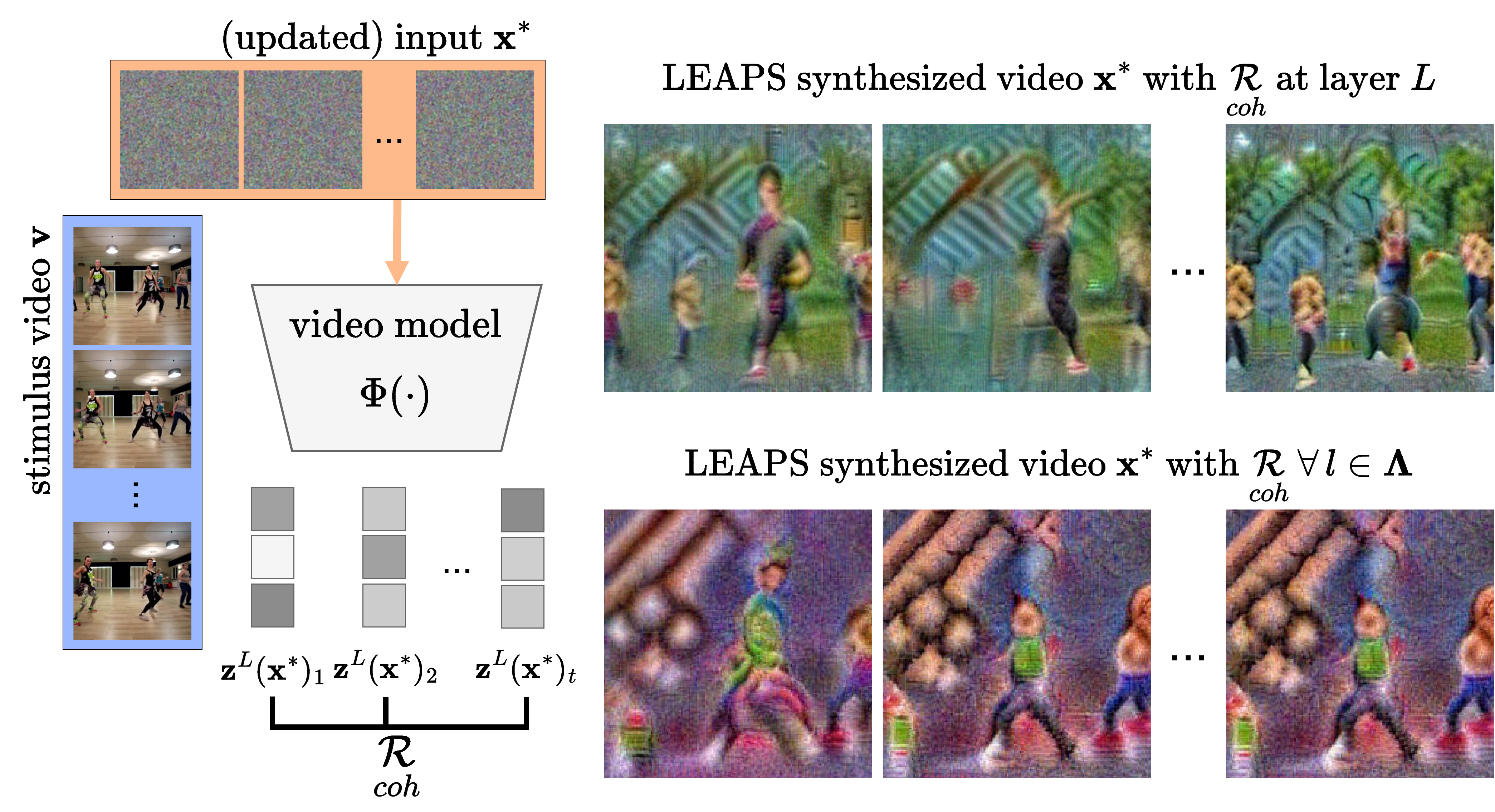}
    \caption{\textbf{Temporal Coherence regularization} of \emph{zumba} target action label. Applying temporal coherence regularization on representations from all layers $l \in \mathbf{\Lambda}$ synthesizes static videos (bottom row). Instead, regularizing only the final network layer $L$ synthesizes videos with consistent frame transitions and motions (top row). We note that the video stimulus is shown as a reference as the regularizations are only applied to synthesized video representations.}
    \label{fig:leaps_tempcoh}
\vspace{-1.1em}
\end{figure}

\subsection{Temporal Coherence Regularization}
\label{sec:method::temp_coh}

For the first regularizer, we aim to enforce similarity between representations of consecutive frames in order to enable consistent feature transitions in the synthesized video. Therefore, we include a coherence regularizer $\underset{coh}{\mathcal{R}}$, formulated based on the temporal coherence loss from \cite{mobahi2009deep}. 

\noindent
Given two spatiotemporal representations $\mathbf{z}^{L}(\mathbf{x}^{*})_{t_1}$ and $\mathbf{z}^{L}(\mathbf{x}^{*})_{t_2}$ at layer $L$, for temporal locations $t_1$ and $t_2$, we use their $l_1$ norm to enforce similarity if $t_1$ and $t_2$ are consecutive in the video. In non-consecutive cases, the divergence between $\mathbf{z}^{L}(\mathbf{x}^{*})_{t_1}$ and $\mathbf{z}^{L}(\mathbf{x}^{*})_{t_2}$ should increase. The coherence regularizer is formulated as:  
\begin{equation}\
\label{eq:coherence}
    \underset{coh}{\mathcal{R}}(\mathbf{x}^{*}) \! = \! \begin{cases}
         || \mathbf{z}^{L}(\mathbf{x}^{*})_{t_1} \! - \! \mathbf{z}^{L}(\mathbf{x}^{*})_{t_2} ||_1 , \text{if consecutive}\\
        \text{max}\!\left( 0,\delta \! - \!|| \mathbf{z}^{L}(\mathbf{x}^{*})_{t_1} \! - \! \mathbf{z}^{L}(\mathbf{x}^{*})_{t_2} ||_1 \right) \! , \text{elsewise}
    \end{cases}
\end{equation}
\noindent
where $\delta$ is a margin hyperparameter. Temporal coherence is enforced at layer $L$ as in~\cite{mobahi2009deep}. Although $\underset{coh}{\mathcal{R}}$ can be applied to any layer $l \in \mathbf{\mathbf{\Lambda}}$, in practice, minimizing (\ref{eq:coherence}) for all layers enforces a very strong regularization, producing synthesized videos with minimal to no cross-frame variations as shown in~\Cref{fig:leaps_tempcoh}. We note that for Transformers using patches of $P^3$ resolution, we first reshape $\textbf{z}^{L}(\mathbf{x}^*)$ from $C'P^3 \times \frac{T'H'W'}{P^3}$ to $C'\! \!\times\! T'\! \times \!H'W'$ before calculating (\ref{eq:coherence}).

\subsection{Feature Diversity Regularization}
\label{sec:method::feat_div}

Our second regularization term $\underset{feat}{\mathcal{R}}$ is responsible for improving the diversity of features generated by model priming. Although priming provides a strong signal based on which the input can be updated, the diversity of features is limited compared to observing multiple instances. Thus, class features varying from those in the stimulus, or features not present in the stimulus, may not be explored during optimization. In order to enhance the search space we introduce an additional domain-specific verifier network $\mathcal{S}(\cdot)$. Our goal is to use high- and low-level feature distribution statistics as proposed in~\cite{yin2020dreaming} incorporating the verifier's prior knowledge during optimization.

\begin{figure}
    \centering
    \includegraphics[width=\linewidth]{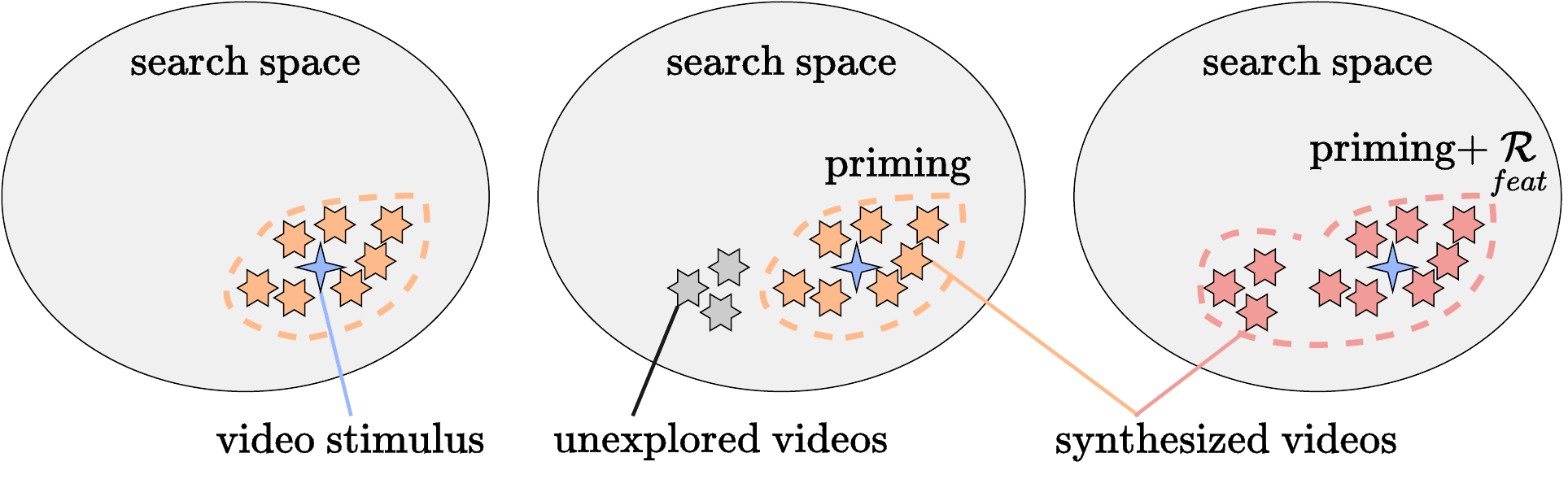}
    \caption{\textbf{Feature diversity regularization}. Given the video stimulus in \textcolor[HTML]{4B67A6}{blue}, synthesized videos in \textcolor[HTML]{FFBA8C}{orange} are optimized only from the stimulus. Synthesized videos with feature diversity regularization are shown in \textcolor[HTML]{DB5C57}{red}. The regularized videos can also include general class features different from or not existing in the stimulus.}
    \label{fig:leaps_featdiv}
\vspace{-1.1em}
\end{figure}

\noindent
 Based on input $\mathbf{x}^{*}$, we run inference on the verifier $\mathcal{S}(\mathbf{x}^{*})$ to obtain representations $\mathbf{a}^{k}(\mathbf{x}^{*})$ across each verifier layer $k \in \textbf{K}$. The feature statistics are then obtained by the $C$-length space-time mean $\mu \left( \textbf{a}^{k}(\mathbf{x}^{*}) \right)$ and variance $\sigma^{2} \left( \textbf{a}^{k}(\mathbf{x}^{*}) \right)$ vectors. The feature diversity regularizer is defined as:
 \begin{equation}
 \label{eq:feat}
\begin{split}
\underset{feat}{\mathcal{R}}(\mathbf{x}^{*}) \! = \! \underset{k \in \mathbf{K}}{\sum}||\mu \left( \textbf{a}^{k}(\mathbf{x}^{*}) \right) - \mathbb{E}\bigl(\mu \left( \textbf{a}^{k}(\mathbf{x}) \right)|\mathcal{X}\bigr)||_2 + \\
\underset{k \in \mathbf{K}}{\sum}||\sigma^{2} \left( \textbf{a}^{k}(\mathbf{x}^{*}) \right) - \mathbb{E} \bigl(\sigma^{2} \left( \textbf{a}^{k}(\mathbf{x}) \right)|\mathcal{X}\bigr)||_2  
\end{split}
\end{equation}
 \noindent
 where $\mathbb{E}(\cdot)$ corresponds to the expected value of representation $\mathbf{a}(\cdot)$ for video input $\mathbf{x}$ part of video dataset $\mathcal{X}$. Instead of requiring access to the dataset to train over $\mathbf{x} \in \mathcal{X}$ videos, we use the Batch Normalization~\cite{ioffe2015batch} running mean and variance to approximate the expected mean and variance as in~\cite{yin2020dreaming}. The mean and variance estimates are then reformulated as $\mathbb{E}\bigl(\mu \left( \textbf{a}^{k}(\mathbf{x}) \right)|\mathcal{X}\bigr) \simeq BN^{k}(\text{running\_mean})$ and $\mathbb{E} \bigl(\sigma^{2} \left( \textbf{a}^{k}(\mathbf{x}) \right)|\mathcal{X}\bigr) \simeq BN^{k}(\text{running\_variance})$. An illustration of the improved search space achieved with the inclusion of feature diversity is shown in~\Cref{fig:leaps_featdiv}.

\subsection{Aggregation for Model Inversion}
\label{sec:method::aggregation}

We have introduced model priming as a method to obtain a strong signal from a stimulus video with which noise-initialized videos can be optimized. 
To implicitly enforce transition continuity across frames, we include a temporal coherence regularizer. In addition, as the stimulus does not provide prior intuition about the diversity of features, we use a feature diversity regularizer to enhance the search space. Given their complementary properties, we formulate the final LEAPS objective as the combination of model priming, temporal coherence, and feature diversity regularizers. In line with spatial feature visualization losses that synthesize distinguishable features from noise~\cite{alexander2015inceptionism,mobahi2009deep,santurkar2019image,yin2020dreaming}, we include a cross-entropy loss $\underset{CE}{\mathcal{L}}$ from class predictions given the synthesized input $\mathbf{x}^{*}$:  
\begin{equation}
\label{eq:leaps}
    \mathcal{L}(\mathbf{x}^{*},\mathbf{v}, y) \! = \! \underset{CE}{\mathcal{L}}(\mathbf{x}^{*},y) + \underset{prim}{\mathcal{L}}(\mathbf{x}^{*},\mathbf{v}) + r \mathcal{R}(\mathbf{x}^{*})
\end{equation}
\noindent
where the regularizer term combines (\ref{eq:coherence}) and (\ref{eq:feat}): $\mathcal{R}(\mathbf{x}^{*}) = \underset{coh}{\mathcal{R}} + \underset{feat}{\mathcal{R}}$. Respectively, $r$ is a regularizer scaling factor. The final LEAPS objective enables the synthesis of high-fidelity features without being constrained by data availability or architecture types as shown in the results section.

\section{Main Results}
\label{sec:results}

We first detail the train scheme and implementation settings used in \Cref{sec:results::details}. We then compare our proposed LEAPS method to image-based feature visualization methods that we extend to video and use as baselines in \Cref{sec:results::comparisons}. We also qualitatively and quantitatively investigate the synthesized videos across a variety of video models in \Cref{sec:results::models}. Finally, in \Cref{sec:results::ablations} we perform ablation studies for the LEAPS objective as well as different updatable input spatiotemporal resolutions.

\subsection{Experimental details}
\label{sec:results::details}

\noindent
\textbf{Model details}. We invert 3D~\cite{hara2018can}/(2+1)D~\cite{tran2018closer}/CSN~\cite{tran2019video} ResNet-50 (R50), X3D~\cite{feichtenhofer2020x3d}, TimeSformer~\cite{bertasius2021space}, Video Swin~\cite{liu2022video}, MViTv2~\cite{li2022mvitv2}, rev-MViT~\cite{mangalam2022reversible}, and UniFormerv2~\cite{li2022uniformerv2} networks. For all our experiments, we use the official networks available from their respective repositories pretrained on Kinetics-400~\cite{carreira2017quo}. Due to its limited computational overhead and requirement of BN layers by the feature diversity regularizer in (\ref{eq:feat}), we use X3D$_{\text{S}}$ as the verifier network $\mathcal{S}(\cdot)$. We note that only the internal representations and predictions from the inverted and verifier models are used. Both models only run inference and remain fixed throughout the feature synthesis.  

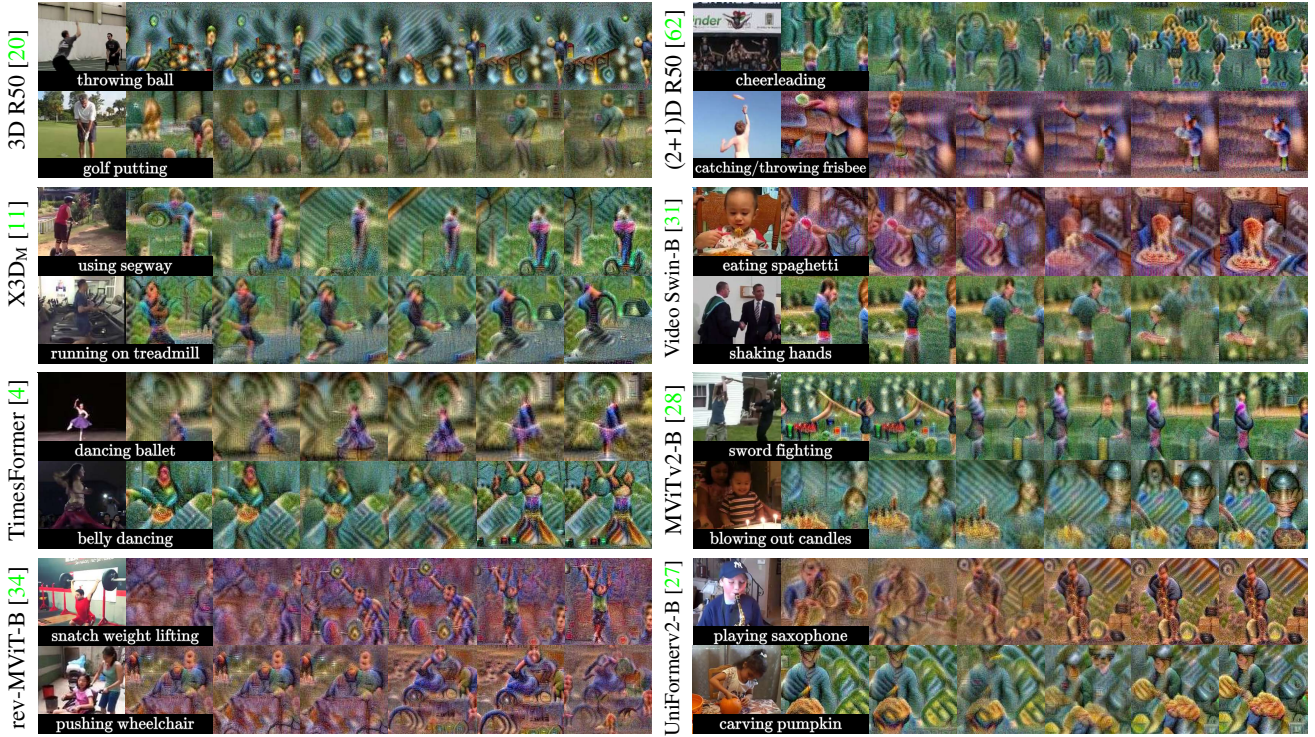
\begin{figure*}[t]
    \centering
    \begin{overpic}[trim = {0cm, 1.5cm, 0cm, 0cm}, clip,width=\textwidth]
    {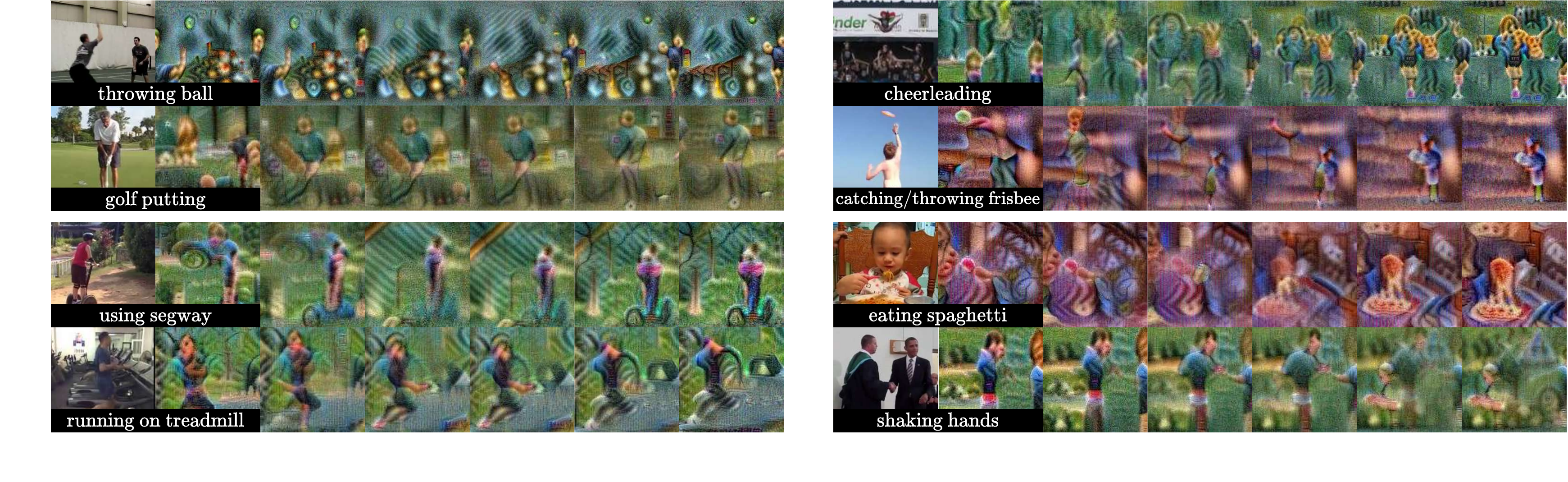}
    \put (1,21) {\rotatebox[origin=c]{90}{\small{3D R50~\cite{hara2018can}}}}
    \put (51,21) {\rotatebox[origin=c]{90}{\small{(2+1)D R50~\cite{tran2018closer}}}}
    \put (1,8.5) {\rotatebox[origin=c]{90}{\small{X3D$_{\text{M}}$~\cite{feichtenhofer2020x3d}}}}
    \put (51,6.7) {\rotatebox[origin=c]{90}{\footnotesize{Video Swin-B~\cite{liu2022video}}}}
    \end{overpic}
    \begin{overpic}[trim = {0cm, .6cm, 0cm, 1.1cm}, clip,width=\textwidth]{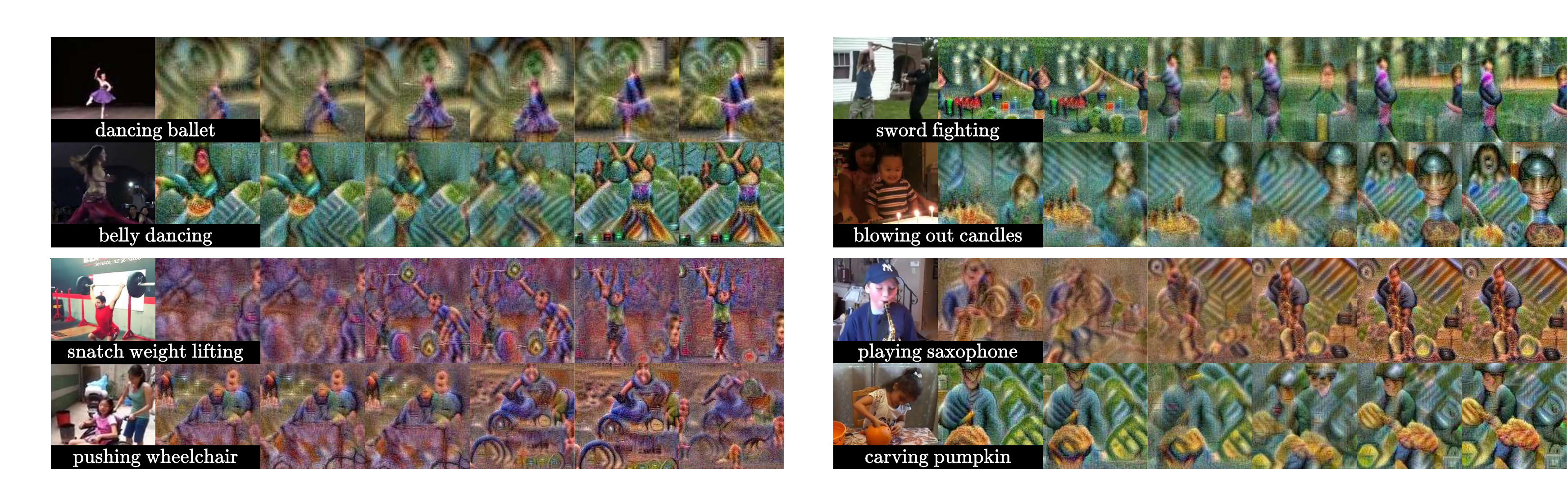}
    \put (1,21) {\rotatebox[origin=c]{90}{\small{TimesFormer~\cite{bertasius2021space}}}}
    \put (51,21) {\rotatebox[origin=c]{90}{\small{MViTv2-B~\cite{li2022mvitv2}}}}
    \put (1,6.7) {\rotatebox[origin=c]{90}{\small{rev-MViT-B~\cite{mangalam2022reversible}}}}
    \put (51,6.7) {\rotatebox[origin=c]{90}{\footnotesize{UniFormerv2-B~\cite{li2022uniformerv2}}}}
    \end{overpic}
    \caption{\textbf{Qualitative examples of synthesized spatiotemporal features with LEAPS}. Models are primed with a stimulus video, shown on the left of each row of synthesized videos.}
    \label{fig:leaps_examples}
\vspace{-1.1em}
\end{figure*}

\begin{figure}[t]
     \centering
     \begin{subfigure}[b]{\linewidth}
         \centering
         \includegraphics[width=\textwidth]{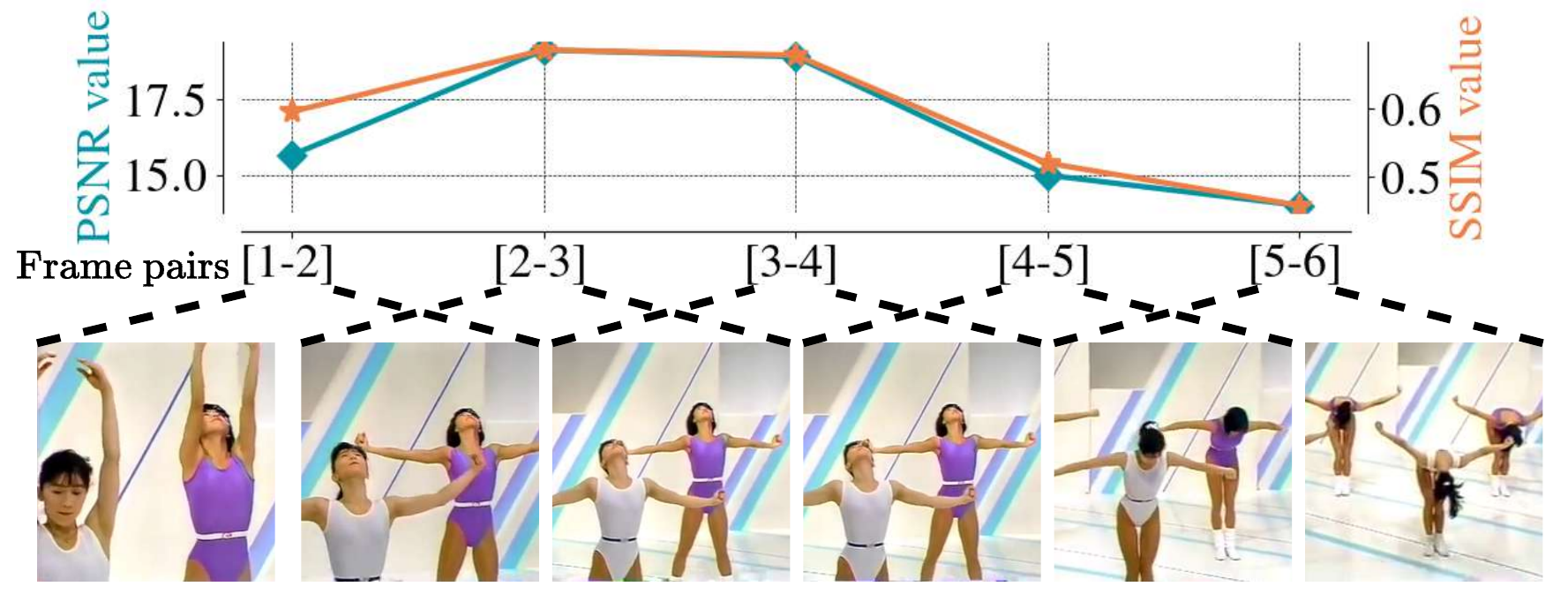}
         \vspace{-1.8em}
         \caption{stimulus video for \textit{doing aerobics}}
         \label{fig:temporal_coh::video}
     \end{subfigure}
     \\
     \begin{subfigure}[b]{\linewidth}
         \centering
         \includegraphics[width=\textwidth]{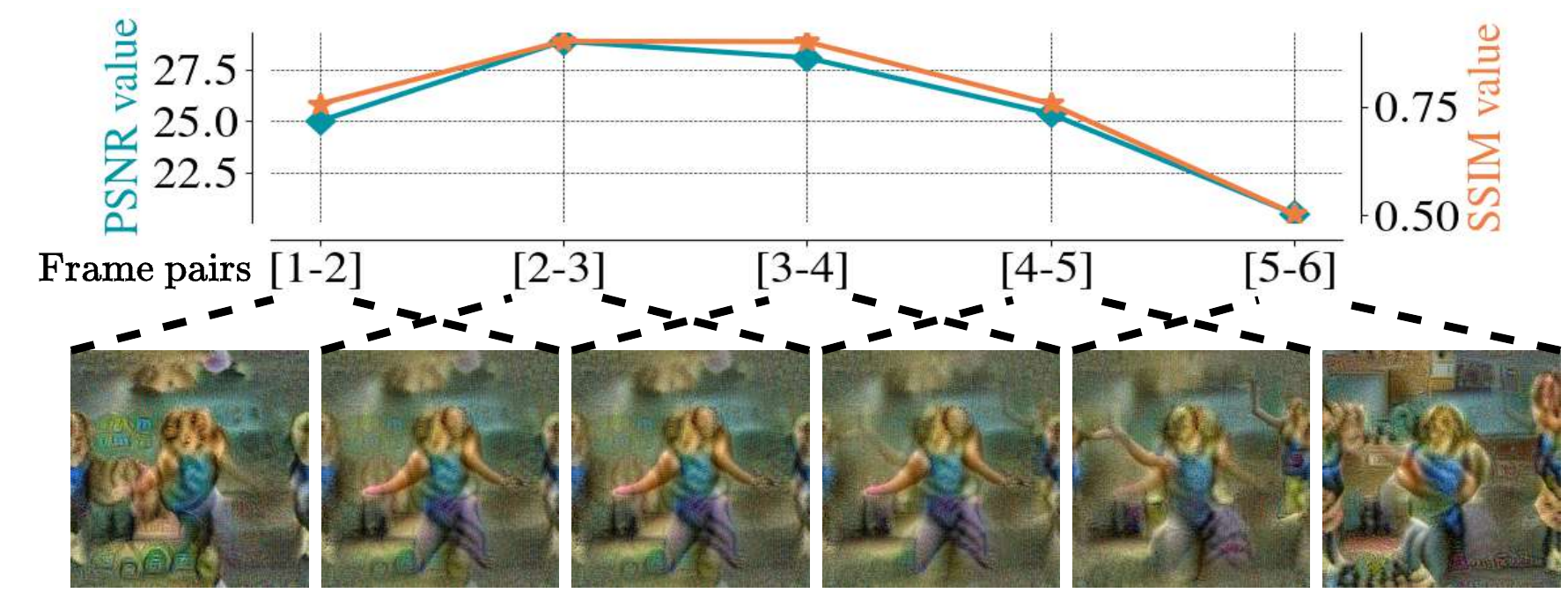}
         \vspace{-1.8em}
         \caption{LEAPS synthesized video}
         \label{fig:temporal_coh::synth}
     \end{subfigure}
     \vspace{-1em}
        \caption{\textbf{SSIM and PSNR statistics over real and synthesized videos} using Video Swin-B features. Lower values correspond to larger cross-frame differences. }
        \label{fig:temporal_coh}
\vspace{-1.1em}
\end{figure}

\noindent
\textbf{Feature synthesis optimization details}. Video input $\mathbf{x}^{*}$ is initialized with size of $8 \times 224^2$. For models~\cite{feichtenhofer2020x3d,bertasius2021space,liu2022video,li2022mvitv2,mangalam2022reversible,li2022uniformerv2} which use inputs of fixed size, we first interpolate $\mathbf{x}^{*}$ to match the required size\footnote{Due to space limitations in \Cref{fig:feat_div,fig:leaps_examples,fig:temporal_coh,fig:iterations} the last 2 frames of $\mathbf{x}^{*}$ are not shown.}. Priming stimuli are selected from the Kinetics-400 validation set randomly. For transformer models, $\mathbf{x}^{*}$ and  $\mathbf{v}$ are first tokenized. We use Adam~\cite{kingma2015adam} with a learning rate of 0.2 and a cosine decrease policy as in~\cite{yin2020dreaming}. We use a total of 2K gradient updates. As in \cite{mobahi2009deep}, we set $\delta=1$. To discover the optimal $\lambda$ and $r$ hyperparameters for each network we use Mango~\cite{sandha2020mango} to perform a simple grid search for 1K gradient updates. A full overview of the hyperparameters used by each model is available in Table~\textcolor{red}{S1} in the supplementary material.

\subsection{Baseline results}
\label{sec:results::comparisons}

We first assess the quality of the synthesized videos. We invert 3D R50, X3D$_{\text{M}}$, and Video Swin-B models and report the averaged top-1 classification accuracies and Inception Scores (IS)~\cite{salimans2016improved} on synthesized videos using each video from the validation set of Kinetics-400 as a stimulus. We extend two prominent image-based feature visualization methods making them applicable to spatiotemporal models:

\noindent
\textbf{DeepDream}~\cite{alexander2015inceptionism} optimizes the input by a cross-entropy loss. It uses two regularizers including the total variance and $\mathit{l}_2$ norm on the input to improve convergence.

\noindent
\textbf{Activation Maximization (AM)}~\cite{mahendran2016visualizing} optimizes a random noise image by gradient ascent to maximize the activation of a specific class. We specifically adapt~\cite{feichtenhofer2020deep} for visualizing concurrent spatiotemporal representations, as it is the only prior method for visualizing features over space and time.

\begin{table}[t]
\centering
\resizebox{\linewidth}{!}{%
\begin{tabular}{ l| c c | l l }
\hline
\multicolumn{1}{c|}{\multirow{2}{*}{Visualization method}} &
\multicolumn{2}{c|}{top-1 (\%)} &
\multicolumn{2}{c}{Inception Score (IS)} \\
& model & ver. & model & verifier \\
\hline
\rowcolor{LightGrey} \multicolumn{5}{l}{\textit{3D R50}} \tstrut \\
3D Deep Dream \cite{alexander2015inceptionism} & 34.5 & 2.6 & 1.1 $\pm$ 0.1 & 1.0 \\[.4em]
3D AM \cite{feichtenhofer2020deep} & 41.4 & 5.8 & 1.4 $\pm$ 0.3 & 1.2 $\pm$ 0.2 \\[.4em]
LEAPS \textbf{ours} ($\underset{prim}{\mathcal{L}}$) & 67.9 & 53.1 & 3.9 $\pm$ 0.9 & 3.2 $\pm$ 0.6 \\
LEAPS \textbf{ours} ($\underset{prim}{\mathcal{L}}$+$\underset{feat}{\mathcal{R}}$) & 74.3 & 60.2  & 5.1 $\pm$ 0.8 & 3.9 $\pm$ 1.4 \\
LEAPS \textbf{ours} \textbf{(full)} & \underline{86.7} & \underline{68.5} & \underline{9.0 $\pm$ 1.0} & \underline{5.7 $\pm$ 0.7} \\
\hline
\rowcolor{LightGrey} \multicolumn{5}{l}{\textit{X3D$_{\text{M}}$}} \tstrut \\
3D Deep Dream \cite{alexander2015inceptionism} & 18.1 & 1.9 & 1.1 $\pm$ 0.1 & 1.1 $\pm$ 0.1 \\[.4em]
3D AM \cite{feichtenhofer2020deep} & 33.2 & 5.6 & 1.3 $\pm$ 0.3 & 1.2 $\pm$ 0.2 \\[.4em]
LEAPS \textbf{ours} ($\underset{prim}{\mathcal{L}}$) & 73.4 & 59.8 & 5.1 $\pm$ 1.1 & 3.9 $\pm$ 0.4 \\
LEAPS \textbf{ours} ($\underset{prim}{\mathcal{L}}$+$\underset{feat}{\mathcal{R}}$) & 81.1 & 69.3 & 8.8 $\pm$ 1.5 & 6.3 $\pm$ 0.8 \\
LEAPS \textbf{ours} \textbf{(full)} & \underline{\textbf{90.3}} & \underline{\textbf{82.5}} & \underline{\textbf{11.4 $\pm$ 0.9}} & \underline{\textbf{8.0 $\pm$ 1.4}} \\
\hline
\rowcolor{LightGrey} \multicolumn{5}{l}{\textit{Video Swin-B}} \tstrut \\
3D Deep Dream \cite{alexander2015inceptionism} & 15.9 & 1.4 & 1.4 $\pm$ 0.2 & 1.1 $\pm$ 0.1 \\[.4em]
3D AM \cite{feichtenhofer2020deep} & 25.6 & 2.2 & 1.6 $\pm$ 0.4 & 1.1 $\pm$ 0.1 \\[.4em]
LEAPS \textbf{ours} ($\underset{prim}{\mathcal{L}}$) & 71.2 & 58.6 & 4.4 $\pm$ 0.8 & 3.5 $\pm$ 0.6 \\
LEAPS \textbf{ours} ($\underset{prim}{\mathcal{L}}$+$\underset{feat}{\mathcal{R}}$) & 76.0 & 65.4 & 5.3 $\pm$ 1.5 & 4.1 $\pm$ 1.1 \\
LEAPS \textbf{ours} \textbf{(full)} & \underline{87.4} & \underline{74.3} & \underline{9.8 $\pm$ 1.3} & \underline{6.5 $\pm$ 0.9} \\
\end{tabular}
}
\caption{\textbf{Quantitative results for mean top-1 accuracy and Inception Score (IS)}. The best results per metric are in \textbf{bold} and per architecture are \underline{underlined}.}
\label{tab:k400_accuracies}
\vspace{-1.1em}
\end{table}

\noindent
\textbf{Quantitative evaluation}. \Cref{tab:k400_accuracies} shows the average top-1 accuracies and IS obtained by both the inverted models and verifier when inferring synthesized videos. For DeepDream and AM that do not use priming, the statistics are averaged across 10 runs per class. In LEAPS the statistics are calculated using each video in the Kinetics validation set as stimulus. For both measures, LEAPS yields consistently higher accuracies and IS compared to DeepDream and AM. Notably, it significantly improves the verifier accuracy across all three architectures. This demonstrates the merits of model priming as both DeepDream and AM optimize inputs solely by maximizing feature or class activations, without using the information-rich internal representations provided by a stimulus. This trend is also visible in the IS, as LEAPS regularizes the synthesized features in order to better represent learned temporally coherent motions.

\begin{figure}[t]
     \centering
     \begin{subfigure}[b]{0.48\linewidth}
         \centering
         \includegraphics[width=\textwidth]{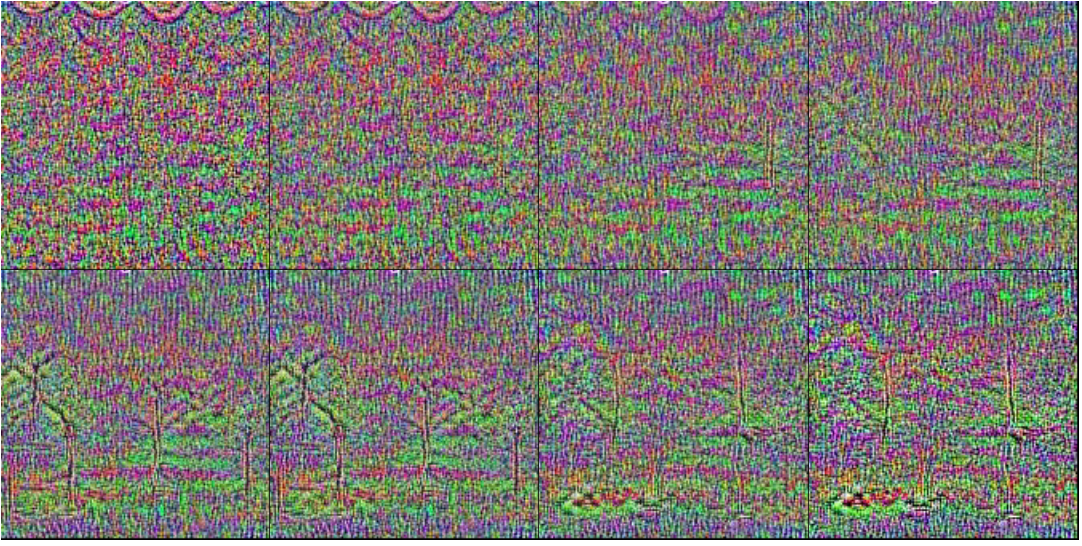}
         \caption{3D DeepDream~\cite{alexander2015inceptionism}}
         \label{fig:comparisons::inceptionism}
     \end{subfigure}
     \hfill
     \begin{subfigure}[b]{0.48\linewidth}
         \centering
         \includegraphics[width=\textwidth]{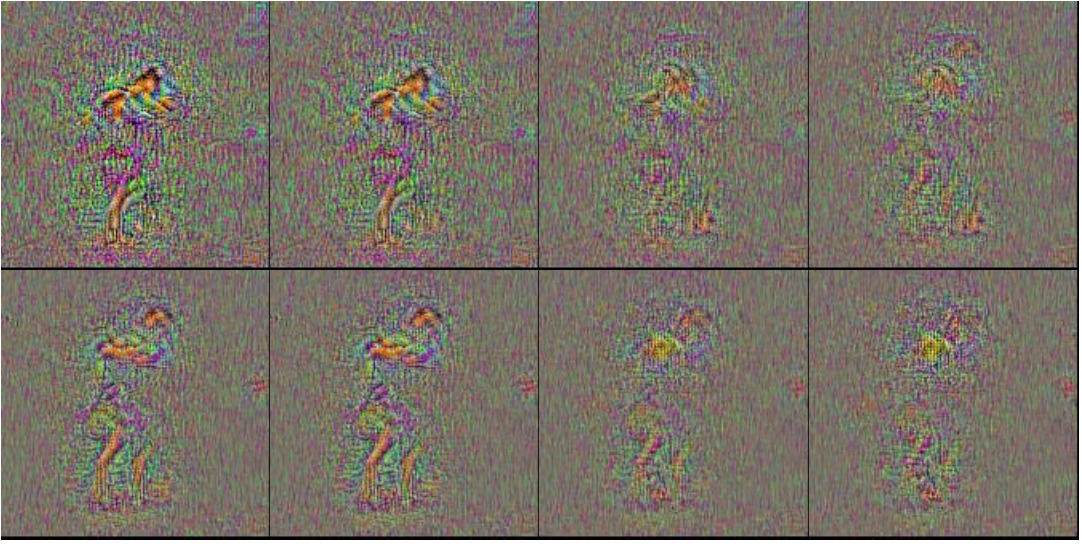}
         \caption{3D AM~\cite{feichtenhofer2020deep}}
         \label{fig:comparisons::am}
     \end{subfigure}
     \put (-125,11) {\scriptsize{Time}}
     \put (-125,0) {\includegraphics[width=.03\textwidth]{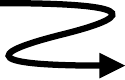}}
     \\[.3em]
     \begin{subfigure}[b]{0.48\linewidth}
         \centering
         \includegraphics[width=\textwidth]{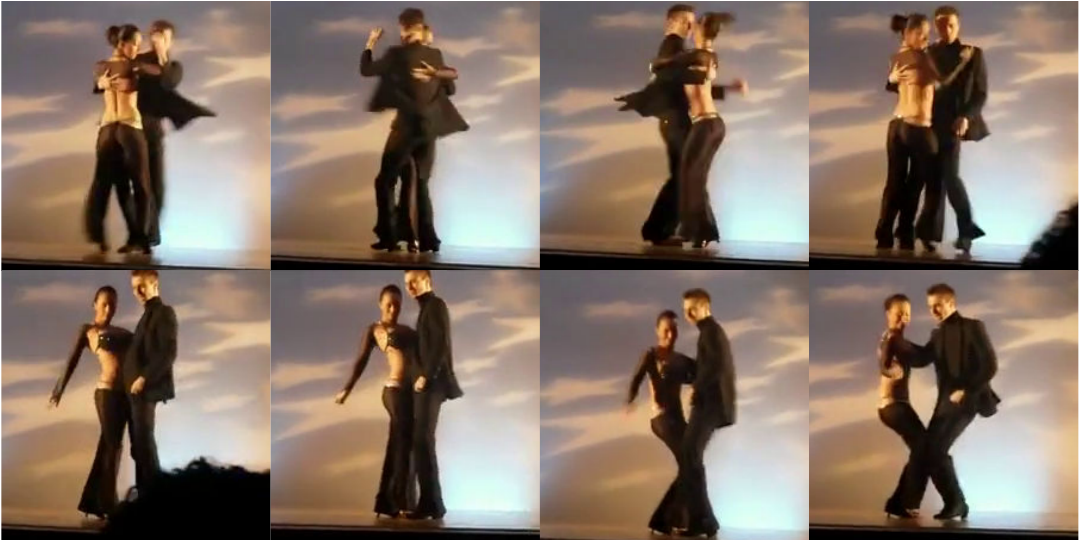}
         \caption{stimulus video}
         \label{fig:comparisons::stimulus}
     \end{subfigure}
     \hfill
     \begin{subfigure}[b]{0.48\linewidth}
         \centering
         \includegraphics[width=\textwidth]{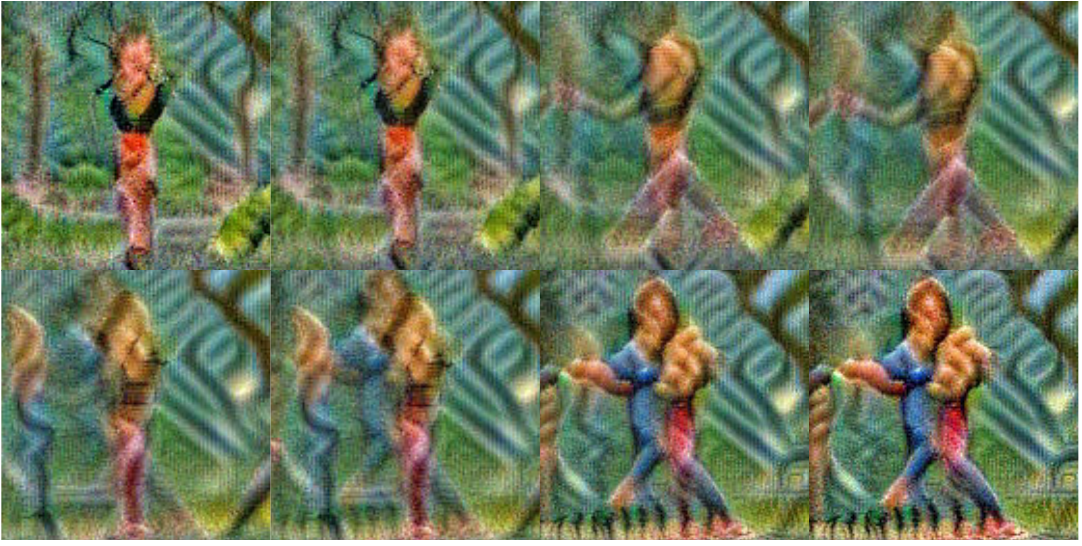}
         \caption{LEAPS}
         \label{fig:comparisons::leaps}
     \end{subfigure}
     \vspace{-.5em}
        \caption{\textbf{Different feature synthesis methods for visualizing X3D$_{\text{M}}$ features} corresponding to class \emph{salsa dancing}. Stimulus video is only used for LEAPS.}
        \label{fig:comparisons}
\end{figure}

\begin{figure}
    \centering
    \begin{overpic}[width=\linewidth]{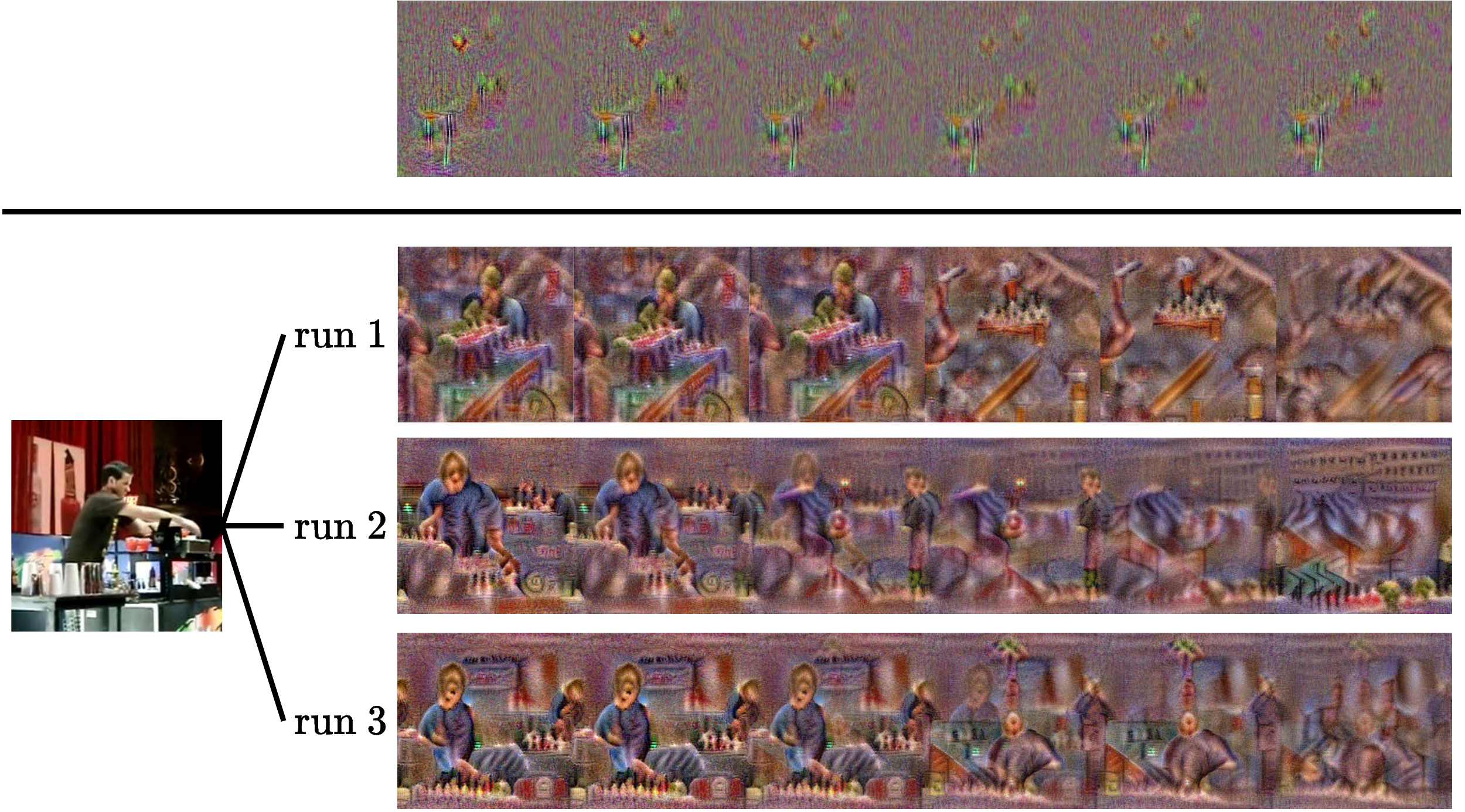}
    \put (1.4,44) {\footnotesize{3D AM~\cite{feichtenhofer2020deep}}}
    \put (1.4,34) {\footnotesize{LEAPS}}
    \put (2,8) {\footnotesize{stimulus}}
    \end{overpic}
    \caption{\textbf{Feature synthesis over different runs} for action class \emph{bartending}. MViTv2 features are from different runs based on the same priming stimulus. }
    \label{fig:feat_div}
\vspace{-1.1em}
\end{figure}

\noindent
\textbf{Qualitative examples}. Videos for the class \emph{salsa spin} synthesized from different methods are shown in~\Cref{fig:comparisons}. Features synthesized with LEAPS are significantly more visually distinct compared to those of baseline methods. As shown, videos produced by image-based methods extended to videos fail to represent learned spatiotemporal deep features. The visual quality of the synthesized videos also correlates with the accuracies and IS in \Cref{tab:k400_accuracies}.

\Cref{fig:feat_div} illustrates synthesized videos from different runs given the same stimulus for class \emph{bartending}. A substantial difference in the quality and representability of the visualizations between LEAPS and the extended 3D AM can be seen across runs. Despite the use of a video stimulus to prime the network, LEAPS visualizations are not constrained by the representations of the stimulus video. Each run of our LEAPS method shows a distinct visual style as feature diversity is encouraged during optimization through the homonym regularizer term. Effectively, LEAPS can be used as a tool for investigating the different class-specific features learned by each model.

\subsection{Analysis of LEAPS synthesized video}
\label{sec:results::models}

The generalizability of feature visualization methods across multiple architectures is largely neglected with features from only a small subset of models being visualized. To evaluate the architecture-independent nature of LEAPS and better understand the visual fidelity of the synthesized videos, we investigate the applicability of LEAPS over a range of convolutional and attention-based video models.   

\noindent
\textbf{Generalizability}. The quality of the produced feature visualizations may depend on the architecture used, as models vary in terms of their complexities and feature spaces that they employ. We compare our proposed LEAPS visualization method by inverting common video models shown in \Cref{fig:leaps_examples} for an arbitrary number of Kinetics classes. A common theme that arises for all models is the association of objects with specific actions. For example, the feature visualizations for the \textit{throwing ball}, \textit{catching/throwing frisbee}, \textit{eating spaghetti}, and \textit{snatch weight lifting} actions from inverted features of 3D R50, (2+1)D R50, Video Swin-B, and rev-MViT-B respectively, all optimize the video to primarily focus on the objects associated with the specific actions. Instead, for actions that are primarily perceived by motions performed, e.g. \textit{running on treadmill} and \textit{dancing ballet}, the actors/performers of the target actions are shown to be more conceptually influential to the model's learned preconscious of an action, with their associated motions and movements captured by the produced visualizations. In addition, the visualizations for cases \textit{eating spaghetti}, \textit{blowing out candles}, and \textit{sword fighting} reveal that networks also learn to temporally bound distinct spatiotemporal features of certain actions. 
This is an important ability to be learned by video models as it effectively demonstrates their capacity to filter temporal information alongside the spatial signal. Notably, LEAPS visualizations for both convolutional and attention-based models show that learned features have a good correspondence to their classes across architectures.      

\noindent
\textbf{Temporal coherence}. As we show in~\Cref{fig:temporal_coh}, LEAPS can visualize motions performed over different speeds (slow/fast). We report the Peak Signal Noise Ratio (PSNR) and Structural Similarity Index Measure (SSIM) for consecutive frame pairs to analyze the variations observed by both fast and slow motions within a video. We observe that the speeds with which motions are performed within the same video can be fundamentally different with both large and small changes occurring between frames. In contrast to image-based methods extended to video, LEAPS shows the ability to represent such variations in the produced feature visualizations. As shown, the PSNR and SSIM statistics from the stimulus video of class \textit{doing aerobics} in~\Cref{fig:temporal_coh::video}, follow similar trends as those produced by the synthesized LEAPS video in~\Cref{fig:temporal_coh::synth}.

\begin{figure}
    \centering
    \begin{overpic}[width=\linewidth]{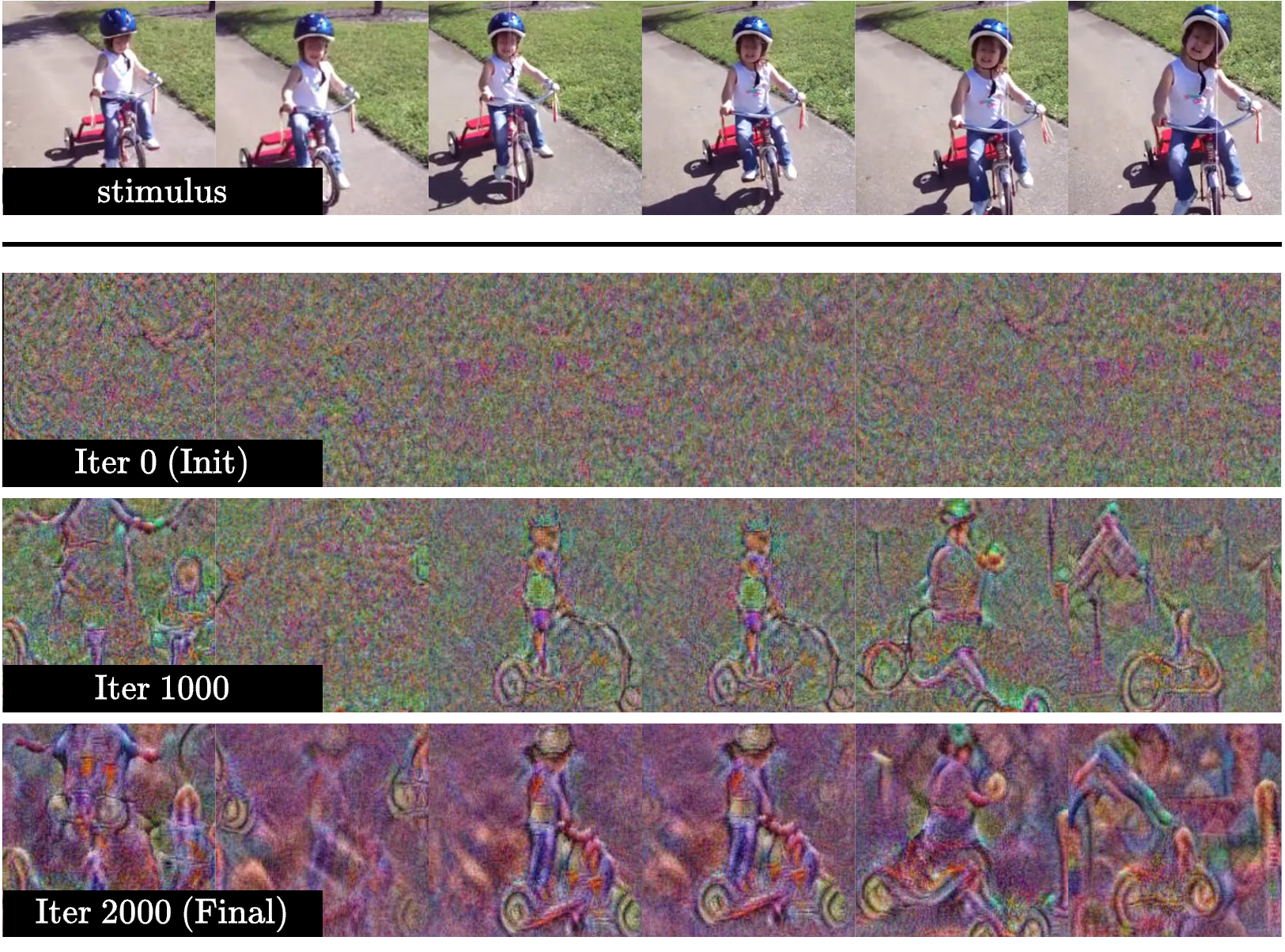}
    \end{overpic}
    \caption{\textbf{Video synthesis at different optimization stages for class} \textit{riding on a bike}. X3D$_{\text{M}}$ features are used.}
    \label{fig:iterations}
\vspace{-1.1em}
\end{figure}

\noindent
\textbf{Video synthesis optimization}. We visualize the resulting synthesized video $\mathbf{x}^*$ at different iteration steps during optimization in~\Cref{fig:iterations}. General features such as the outlines of objects and actors as well as their movements are synthesized first by our proposed method. Interestingly, later iterations show to refine the visualized features by further synthesizing visual details. This demonstrates a level of learned hierarchy by video models as to the types of spatiotemporal learned features that are associated with a specific action. Our proposed use of model priming in tandem with temporal coherence and feature diversity regularization terms shows to enable the visualization of models' spatiotemporal representations of actions, at a finer quality and detail.

\begin{table}[t]
\begin{subtable}[t]{0.48\linewidth}
\vspace{-3pt}
\subcaption{\textbf{3D R50}. }
\label{tab:priming_dist::r3d}
\centering
\resizebox{\textwidth}{!}{%
\begin{tabular}{l| c c c c }
\hline
\multirow{2}{*}{Metric} &
\multicolumn{4}{c}{Distance function} \tstrut \\[.2em]
&
$l_2$ &
$l_1$ &
$cos$ &
JVS \bstrut \\
\hline
top-1 (m) & 84.3 & 82.4 & 72.1 & \textbf{86.7} \tstrut \\
top-1 (v) & 65.8 & 63.7 & 46.1 & \textbf{68.5} \\ 
\end{tabular}
}
\end{subtable}
\hfill
\begin{subtable}[t]{0.48\linewidth}
\vspace{-3pt}
\subcaption{\textbf{X3D$_{\text{M}}$}. }
\label{tab:priming_dist::x3d}
\centering
\resizebox{\textwidth}{!}{%
\begin{tabular}{l| c c c c }
\hline
\multirow{2}{*}{Metric} &
\multicolumn{4}{c}{Distance function} \tstrut \\[.2em]
&
$l_2$ &
$l_1$ &
$cos$ &
JVS \bstrut \\
\hline
top-1 (m) & 88.1 & 86.8 & 78.9 & \textbf{90.3} \tstrut \\
top-1 (v) & 79.7 & 78.4 & 70.2 & \textbf{82.5} \\ 
\end{tabular}
}
\end{subtable}
\\
\begin{subtable}[t]{0.48\linewidth}
\vspace{1pt}
\subcaption{\textbf{Video Swin-B}. }
\label{tab:priming_dist::swin}
\centering
\resizebox{\textwidth}{!}{%
\begin{tabular}{l| c c c c }
\hline
\multirow{2}{*}{Metric} &
\multicolumn{4}{c}{Distance function} \tstrut \\[.2em]
&
$l_2$ &
$l_1$ &
$cos$ &
JVS \bstrut \\
\hline
top-1 (m) & 85.5 & 83.0 & 69.2 & \textbf{87.4} \tstrut \\
top-1 (v) & 72.8 & 71.6 & 41.4 & \textbf{74.3} \\ 
\end{tabular}
}
\end{subtable}
\hfill
\begin{subtable}[t]{0.48\linewidth}
\vspace{1pt}
\subcaption{\textbf{MViTv2-B}. }
\label{tab:priming_dist::mvitv2}
\centering
\resizebox{\textwidth}{!}{%
\begin{tabular}{l| c c c c }
\hline
\multirow{2}{*}{Metric} &
\multicolumn{4}{c}{Distance function} \tstrut \\[.2em]
&
$l_2$ &
$l_1$ &
$cos$ &
JVS \bstrut \\
\hline
top-1 (m) & 83.5 & 81.9 & 70.7 & \textbf{85.9} \tstrut \\
top-1 (v) & 72.4 & 69.7 & 45.6 & \textbf{73.1} \\ 
\end{tabular}
}
\end{subtable}
\caption{\textbf{Top-1 accuracies of the inverted model (m) and verifier (v)} on synthesized videos over different priming distance functions. Best results are in \textbf{bold}.}
\label{tab:priming_dist}
\vspace{-1.1em}
\end{table}

\begin{table*}[t]
\centering
\resizebox{\linewidth}{!}{%
\begin{tabular}{l c c c | c c c c | c | c c c | c c | c | c c }
\hline
\multirow{3}{*}{Metric}&
\multicolumn{16}{c}{Video model architectures and variants} \tstrut \bstrut \\
&
\multicolumn{3}{c|}{R50} &
\multicolumn{4}{c|}{X3D~\cite{feichtenhofer2020x3d}} &
\multicolumn{1}{c|}{\multirow{2}{*}{TS~\cite{bertasius2021space}}} &
\multicolumn{3}{c|}{Video Swin~\cite{liu2022video}} &
\multicolumn{2}{c|}{MViTv2~\cite{li2022mvitv2}} &
\multicolumn{1}{c|}{\multirow{2}{*}{rev-MViT-B~\cite{mangalam2022reversible}}} &
\multicolumn{2}{c}{UniFormerv2~\cite{li2022uniformerv2}} \tstrut \bstrut \\
&
3D &
(2+1)D~\cite{tran2018closer}&
CSN~\cite{tran2019video}&
XS &
S &
M &
L &
&
T &
S &
B &
S &
B &
&
B &
L \tstrut \bstrut
\\
\hline
\rowcolor{LightGrey} \multicolumn{17}{c}{\textbf{LEAPS} $\underset{prim}{\mathcal{L}}$} \tstrut \\
top-1 (m) &
67.9 &
63.8 &
72.4 &
67.4 & 68.5 & 73.4 & 73.8 &
64.9 &
69.1 & 69.5 & 71.2 &
70.6 & 71.5 &
64.5 &
72.3 & 73.0 \\
top-1 (v) &
53.1 &
49.5 &
55.8 &
51.9 & \textcolor{lightgray}{68.5} & 59.8 & 60.4 &
52.6 &
52.3 & 53.5 & 57.6 &
54.6 & 56.7 &
50.7 &
56.4 & 56.9 \\
IS &
$3.9 \! \pm \! 0.9$ &
$2.4 \! \pm \! 0.5$ &
$4.2 \! \pm \! 0.6$ &
$4.0 \! \pm \! 1.0$ & $4.3 \! \pm \! 0.6$ & $5.1 \! \pm \! 1.1$ & $5.5 \! \pm \! 1.3$ &
$2.7 \! \pm \! 0.7$ &
$4.1 \! \pm \! 1.6$ & $4.2 \! \pm \! 1.1$ & $4.4 \! \pm \! 0.8$ &
$4.3 \! \pm \! 1.2$ & $4.6 \! \pm \! 0.7$ &
$3.1 \! \pm \! 1.7$ &
$4.3 \! \pm \! 0.8$ & $4.5 \! \pm \! 1.2$ \bstrut \\
\hline
\rowcolor{LightGrey} \multicolumn{17}{c}{\textbf{LEAPS} $\underset{prim}{\mathcal{L}} + \underset{coh}{\mathcal{R}}$} \tstrut \\
top-1 (m) &
70.4 &
65.7 &
76.1 &
74.8 & 75.6 & 78.0 & 78.5 &
73.1 &
72.8 & 73.2 & 74.5 &
73.7 & 74.2 &
69.2 &
74.9 & 75.3  \\
top-1 (v) &
55.8 &
54.3 &
60.1 &
61.0 & \textcolor{lightgray}{75.6} & 65.6 & 70.0 &
63.4 &
62.9 & 63.3 & 63.8 &
63.5 & 63.9 &
52.9 &
63.7 & 64.5  \\
IS &
$4.6 \! \pm \! 1.2$ &
$3.5 \! \pm \! 0.8$ &
$4.8 \! \pm \! 1.6$ &
$4.3 \! \pm \! 0.9$ & $5.0 \! \pm \! 1.5$ & $7.2 \! \pm \! 1.3$ & $7.5 \! \pm \! 1.0$ &
$3.9 \! \pm \! 0.9$ &
$4.5 \! \pm \! 0.7$ & $4.8 \! \pm \! 1.3$ & $5.4 \! \pm \! 0.6$ &
$5.0 \! \pm \! 0.8$ & $5.3 \! \pm \! 1.0$ &
$3.3 \! \pm \! 2.1$ &
$5.5 \! \pm \! 0.6$ & $5.9 \! \pm \! 0.4$  \bstrut \\
\hline
\rowcolor{LightGrey} \multicolumn{17}{c}{\textbf{LEAPS} $\underset{prim}{\mathcal{L}} + \underset{feat}{\mathcal{R}}$} \tstrut \\
top-1 (m) &
74.3 &
67.4 &
76.2 &
78.9 & 79.4 & 81.1 & 81.5 &
71.9 &
74.1 & 74.7 & 76.0 &
75.6 & 76.3 &
70.4 &
75.8 & 76.6  \\
top-1 (v) &
60.2 &
56.9 &
64.3 &
65.8 & \textcolor{lightgray}{79.4} & 69.3 & 70.8 &
67.2 &
63.2 & 64.5 & 65.4 &
66.7 & 68.0 &
58.2 &
69.0 & 69.8  \\
IS &
$5.1 \! \pm \! 0.8$ &
$4.2 \! \pm \! 1.4$ &
$5.6 \! \pm \! 1.2$ &
$7.3 \! \pm \! 1.1$ & $7.6 \! \pm \! 0.8$ & $8.8 \! \pm \! 1.5$ & $9.1 \! \pm \! 1.2$ &
$3.8 \! \pm \! 1.2$ &
$4.7 \! \pm \! 1.1$ & $4.9 \! \pm \! 0.5$ & $5.3 \! \pm \! 1.5$ &
$5.1 \! \pm \! 0.6$ & $5.5 \! \pm \! 1.2$ &
$4.0 \! \pm \! 1.8$ &
$5.7 \! \pm \! 1.0$ & $6.2 \! \pm \! 0.9$  \bstrut \\
\hline
\rowcolor{LightGrey} \multicolumn{17}{c}{\textbf{LEAPS (full})} \tstrut \\
top-1 (m) &
\textbf{86.7} &
\textbf{78.0} &
\textbf{88.3} &
\textbf{86.2} & \textbf{87.0} & \textbf{90.3} & \textbf{90.8} &
\textbf{83.6} &
\textbf{85.7} & \textbf{86.2} & \textbf{87.4} &
\textbf{85.1} & \textbf{85.9} &
\textbf{82.5} &
\textbf{87.1} & \textbf{88.3}  \\
top-1 (v) &
\textbf{68.5} &
\textbf{65.2} &
\textbf{71.6} &
\textbf{76.4} & \textcolor{lightgray}{87.0} & \textbf{82.5} & \textbf{83.7} &
\textbf{69.7} &
\textbf{71.9} & \textbf{73.5} & \textbf{74.3} &
\textbf{72.4} & \textbf{73.1} &
\textbf{67.3} &
\textbf{75.4} & \textbf{76.2}  \\
IS &
$\textbf{9.0} \! \pm \! \textbf{1.0}$ &
$\textbf{6.4} \! \pm \! \textbf{1.3}$ &
$\textbf{9.7} \! \pm \! \textbf{0.7}$ &
$\textbf{9.6} \! \pm \! \textbf{0.4}$ & $\textbf{10.4} \! \pm \! \textbf{1.2}$ & $\textbf{11.4} \! \pm \! \textbf{0.9}$ & $\textbf{11.9} \! \pm \! \textbf{1.5}$ &
$\textbf{7.5} \! \pm \! \textbf{1.6}$ &
$\textbf{8.5} \! \pm \! \textbf{0.7}$ & $\textbf{9.4} \! \pm \! \textbf{1.5}$ & $\textbf{9.8} \! \pm \! \textbf{1.3}$ &
$\textbf{8.7} \! \pm \! \textbf{1.3}$ & $\textbf{9.3} \! \pm \! \textbf{0.8}$ &
$\textbf{7.1} \! \pm \! \textbf{2.6}$ &
$\textbf{9.1} \! \pm \! \textbf{1.3}$ & $\textbf{9.6} \! \pm \! \textbf{1.2}$  \\
\end{tabular}
}
\caption{\textbf{Top-1 accuracies and Inceprion Scores over different objectives} across different architectures and model variants. X3D$_{\text{S}}$ is used as the verifier (v) for all experiments. The best results per variant are in \textbf{bold}. The verifier accuracy is denoted in \textcolor{lightgray}{gray} for the case of using X3D$_{\text{S}}$ for both model inversion and as the verifier.}
\label{tab:objective}
\vspace{-1.1em}
\end{table*}

\begin{table}[t]
\vspace{-3pt}
\vspace{7.5pt}
\centering
\resizebox{\linewidth}{!}{%
\begin{tabular}{ c| c | c c | l | l }
\hline
\multirow{2}{*}{Model} & \multirow{2}{*}{temp.$\times$spatial$^2$} & \multicolumn{2}{c|}{top-1} & \multicolumn{1}{c|}{\multirow{2}{*}{IS}} & Latency (secs) \tstrut \\
& & m & v & & \multicolumn{1}{c}{($\downarrow$I /\ $\uparrow$ B)} \bstrut \\
\hline
\multirow{3}{*}{3D R50} & $8 \times 182^{2}$ & 84.3 & 67.2 & $8.1 \! \pm \! 1.3$ & \textbf{0.591 / 0.913} \tstrut \\
& $8 \times 224^{2}$ & 86.7 & 68.5 & $9.7 \! \pm \! 1.0$ & 0.832 / 1.205 \\
& $16 \times 224^{2}$ & \textbf{87.0} & \textbf{68.7} & $\textbf{9.8} \! \pm \! \textbf{1.6}$ & 1.140 / 1.681 \bstrut \\
\hline
\multirow{3}{*}{(2+1)D R50} & $8 \times 182^{2}$ & 75.4 & 62.9 & $4.2 \! \pm \! 1.8$ & \textbf{1.684/2.325} \tstrut \\
& $8 \times 224^{2}$ & 78.0 & 65.2 & $6.4 \! \pm \! 1.3$ & 1.858/2.793 \\
& $16 \times 224^{2}$ & \textbf{78.5} & \textbf{65.4} & $\textbf{6.6} \! \pm \! \textbf{1.5}$ & 2.343/3.176 \\

\end{tabular}
}
\caption{\textbf{Synthesized video size comparisons} based on the top-1 accuracies of the inverted model, and verifier, Inception Scores (IS), and latency times for inference ($\downarrow$I) and backprop ($\uparrow$B). Best settings per architecture are in \textbf{bold}.}
\label{tab:video_sizes}
\vspace{-1.1em}
\end{table}

\subsection{Ablation studies}
\label{sec:results::ablations}

In this section, we conduct ablation studies reporting model statistics over synthesized videos. We initially consider the effect of the distance function used during model priming. Additionally, we report statistics over different combinations of priming and introduced regularizers across the tested architectures. Finally, we present quantitative results when using inputs of different spatiotemporal sizes alongside the resulting averaged latency times.

\noindent
\textbf{Priming distance methods}. In~\Cref{tab:priming_dist}, we evaluate the impact of the distance functions used during model priming at (\ref{eq:priming}). We test three different magnitude-based methods including $l_2$, $l_1$, and $JVS$, as well the cosine similarity $cos$. Across all four spatiotemporal models, 3D R50, X3D$_{\text{M}}$, Video Swin-B, and MViTv2-B, the magnitude-based methods perform favorably over the cosine similarity. This is due to $cos$ not taking into account the magnitude of the feature activation vectors from the synthesized and stimulus videos, which in turn limits the ability to synthesize representations relevant to the stimulus. Across magnitude-based methods, an average improvement of $+2.2\%$ and $+4.1\%$ to the inverted model's (m) accuracy is observed with JVS compared to $l_2$ and $l_1$ respectively. The same trend also holds true for the verifier (v), with $+1.9\%$ and $+3.7\%$ accuracy improvements from $l_2$ and $l_1$ when using JVS. Comparatively to $l_2$ and $l_1$, JVS uses a combination of vector magnitudes and angles~\cite{fernando2021anticipating}, providing a more balanced approach than magnitude-only or angle-only metrics. Therefore, we adopt $JVS$ as the priming distance method used between stimulus and synthesized feature vectors.

\noindent
\textbf{Regularizers}. In~\Cref{tab:objective} we provide comparisons over different priming and regularizer combinations for our LEAPS objective. We report the top-1 accuracies of the inverted model (m), and verifier (v), as well as the IS of the inverted model across a range of spatiotemporal convolutional and attention-based architectures. We note that in the case of inverting X3D$_{\text{S}}$ the same model is used for both model inversion and as the verifier, evidently resulting in matching verifier/inverted model accuracies. Overall, the priming objective $\underset{prim}{\mathcal{L}}$ combined with either regularizer term yields clear improvements compared to the sole use of model priming. Modest improvements are observed with the combination of priming and the feature diversity regularizer over priming with temporal coherence. We believe that this is due to the enhanced search space of $\underset{prim}{\mathcal{L}}+\underset{feat}{\mathcal{R}}$ as more diverse class features not in the stimulus are also explored. The combination of the priming objective with both regularizer terms for our proposed LEAPS consistently achieves the best results by a large margin compared to all other settings. LEAPS improves accuracy, for both the inverted model and verifier, as well as the quality of the generated videos based on the IS. These results further demonstrate that our proposed LEAPS optimization can be used across a range of architectures as a general spatiotemporal feature visualization method. Further qualitative examples for each network over different Kinetics classes are shown in Figures~\textcolor{red}{S1} to~\textcolor{red}{S4} in the supplementary alongside embeddings projections of LEAPS with different regularizer regimes  in Section~\textcolor{red}{S3}.  

\noindent
\textbf{Synthesized video resolution}. Finally, we compare the accuracies, IS, and latency times when optimizing video inputs of different spatiotemporal sizes. Due to the majority of architectures requiring fixed-size inputs, we use 3D R50 and (2+1)D R50 as they are input size independent. From the top-1 (m/v) accuracies and IS summarized in~\Cref{tab:video_sizes}, the best-performing setting across metrics is obtained with $16 \! \times \! 224^2$-sized inputs. We observe comparable performance on the temporally-reduced $8 \! \times \! 224^2$ setting with a significantly more balanced performance-to-latency. Notable decreases in accuracies and IS are observed with the potentially limited spatial resolution $8 \! \times \! 182^2$ setting, making it less suitable. Due to the significant improvements in the per-iteration latency of the $8 \! \times \! 224^2$ inputs, we adopt this setting throughout our experiments.

\section{Conclusions}

We have introduced LEAPS, a novel spatiotemporal model inversion method for visualizing the learned internal representations of networks through video synthesis. LEAPS uses a stimulus video to prime a model and iteratively optimize an input by minimizing the classification and priming loss. The resulting synthesized video visualizes learned concepts that are associated with classes without prior knowledge of the training data. During optimization, LEAPS uses two regularizers. The first enforces temporal coherence between feature transitions across frames in the updatable input. The second improves the diversity of the synthesized features through a domain-specific verifier network to enhance the search space. The proposed architecture-independent method has shown qualitatively and quantitatively that it can produce high-quality and visually coherent synthesized videos over a wide range of spatiotemporal convolutional and attention-based video models. The high classification scores and synthesized video quality metrics make LEAPS a generalizable and effective spatiotemporal feature visualization method. We believe that this first step towards learned spatiotemporal representations synthesis is a promising direction in understanding video models.

\noindent
\textbf{Acknowledgments}. We use publicly available datasets and models. Research is funded by imec.icon Surv-AI-llance project and FWO (Grant G0A4720N).

{\small
\bibliographystyle{ieee_fullname}
\bibliography{egbib}
}

\clearpage

\setcounter{section}{0}
\setcounter{equation}{0}
\setcounter{figure}{0}
\setcounter{table}{0}
\SupplementaryMaterials

\twocolumn[{%
\begin{center}
    \centering
    \textbf{\Large{Leaping Into Memories: Space-Time Deep Feature Synthesis \\ Supplementary material}}\\
    \hspace{2em}
    \vspace{3em}
\end{center}%
}]

\section{Additional Qualitative results}

We demonstrated and overviewed qualitative results of inverted spatiotemporal models with LEAPS in
Section~\textcolor{red}{4.2}. Supplementary to Figure~\textcolor{red}{4}, we provide additional results in order to visualize the features synthesized from different encoders when using the same action labels and stimuli. As shown in \Cref{fig:juggling_1,fig:crawling,fig:dribbling,fig:trumpet} LEAPS can synthesize coherent visual features and effectively invert learned representations, independently of the spatiotemporal architecture used. Similar to the synthesized videos in Figure~\textcolor{red}{4}, for actions that are best described by the objects used e.g. \emph{juggling balls}, \emph{dribbling basketball}, and \emph{playing trumpet}, all models optimize the input video to represent both class-relevant objects as well as actor-object interaction. Importantly, the synthesized videos show that video models learn motions with respect to both objects as well as actors. For the synthesized videos of \emph{juggling balls} in \Cref{fig:juggling_1}, the balls are primarily shown to be thrown upwards. In contrast, for the \emph{dribbling basketball} videos in \Cref{fig:dribbling}, basketballs are bouncing on the side of the actor. In addition, evidence of LEAPS's ability to synthesize class-relevant features can be seen in \Cref{fig:trumpet} where for the \emph{playing trumpet} stimulus used, the better half of the trumpet is occluded. In actions that do not include or cannot be associated with specific objects; e.g. \emph{baby crawling} in \Cref{fig:crawling}, the synthesized videos primarily focus on the actor. This demonstrates that learned class-specific concepts of video models can be based on either objects, the actor's appearance and motions, or both, depending on the action performed.

Based on the videos from inverted models presented in \Cref{fig:juggling_1,fig:crawling,fig:dribbling,fig:trumpet} there are no significant differences as to the objects and actors that are synthesized. However, the level of detail in the synthesized videos is shown to correlate with the model complexity. Specifically for \emph{baby crawling} and \emph{playing trumpet} synthesized videos from inverted models of increased capacities; e.g. X3D, Swin, and MViTv2 contain more visually distinct concepts than those of smaller architectures; e.g. 3D/(2+1)D Resnet-50. The effect is in line with the resulting synthesized videos from inverted models in Figure~\textcolor{red}{6}. Overall, LEAPS can invert models of varying complexities while also visualizing feature details based on the model's feature space capacity.  

\begin{table}[t]
    \centering
    \begin{tabular}{l|l c l | c}
        \hline
        Model & \multicolumn{1}{c}{$\lambda_1$} & $\lambda_L$ & \multicolumn{1}{c|}{$r$} & $\mathcal{L}$ \tstrut \bstrut \\
        \hline
         3D R50 & 1.0 & 0.3 & $7.5e^{-3}$ & 7.892 \tstrut \\
         (2+1)D R50 & 0.75 & 0.1 & $5e^{-3}$ & 6.421 \\ 
         CSN R50 & 1.0 & 0.2 & $5e^{-3}$ & 6.603 \\ 
         X3D$_{\text{XS}}$ & 1.0 & 0.2 & $1e^{-3}$ & 5.175 \\
         X3D$_{\text{S}}$ & 1.0 & 0.1 & $1e^{-3}$ & 5.538 \\ 
         X3D$_{\text{M}}$ & 0.75 & 0.1 & $1e^{-3}$ & 6.387 \\ 
         X3D$_{\text{L}}$ & 0.75 & 0.1 & $1e^{-3}$ & 7.190 \\ 
         TimeSformer & 1.0 & 0.2 & $2.5e^{-3}$ & 5.629 \\ 
         Video Swin-T & 0.75 & 0.2 & $1e^{-3}$ & 6.527 \\
         Video Swin-S & 0.75 & 0.1 & $1e^{-3}$ & 7.508 \\ 
         Video Swin-B & 0.625 & 0.1 & $1e^{-3}$ & 8.841 \\ 
         MViTv2-S & 0.75 & 0.1 & $2.5e^{-3}$ & 7.356 \\ 
         MViTv2-B & 0.75 & 0.1 & $1e^{-3}$ & 8.195 \\ 
         rev-MViT-B & 0.625 & 0.1 & $5e^{-3}$ & 7.227 \\ 
         UniFormerv2-B & 1.0 & 0.2 & $2.5e^{-3}$ & 6.053 \\ 
         UniFormerv2-L & 1.0 & 0.1 & $1e^{-3}$ & 7.415 \\ 
    \end{tabular}
    \caption{\textbf{LEAPS optimization hyperparameters} based on grid search. We additionally report the average loss on synthesized videos from the Kinetics validation set.}
    \label{tab:hyperparams}
\end{table}

\begin{figure*}[t]
    \centering
    \begin{overpic}[trim = {0cm, 0cm, 0cm, 0cm}, clip,width=\textwidth]
    {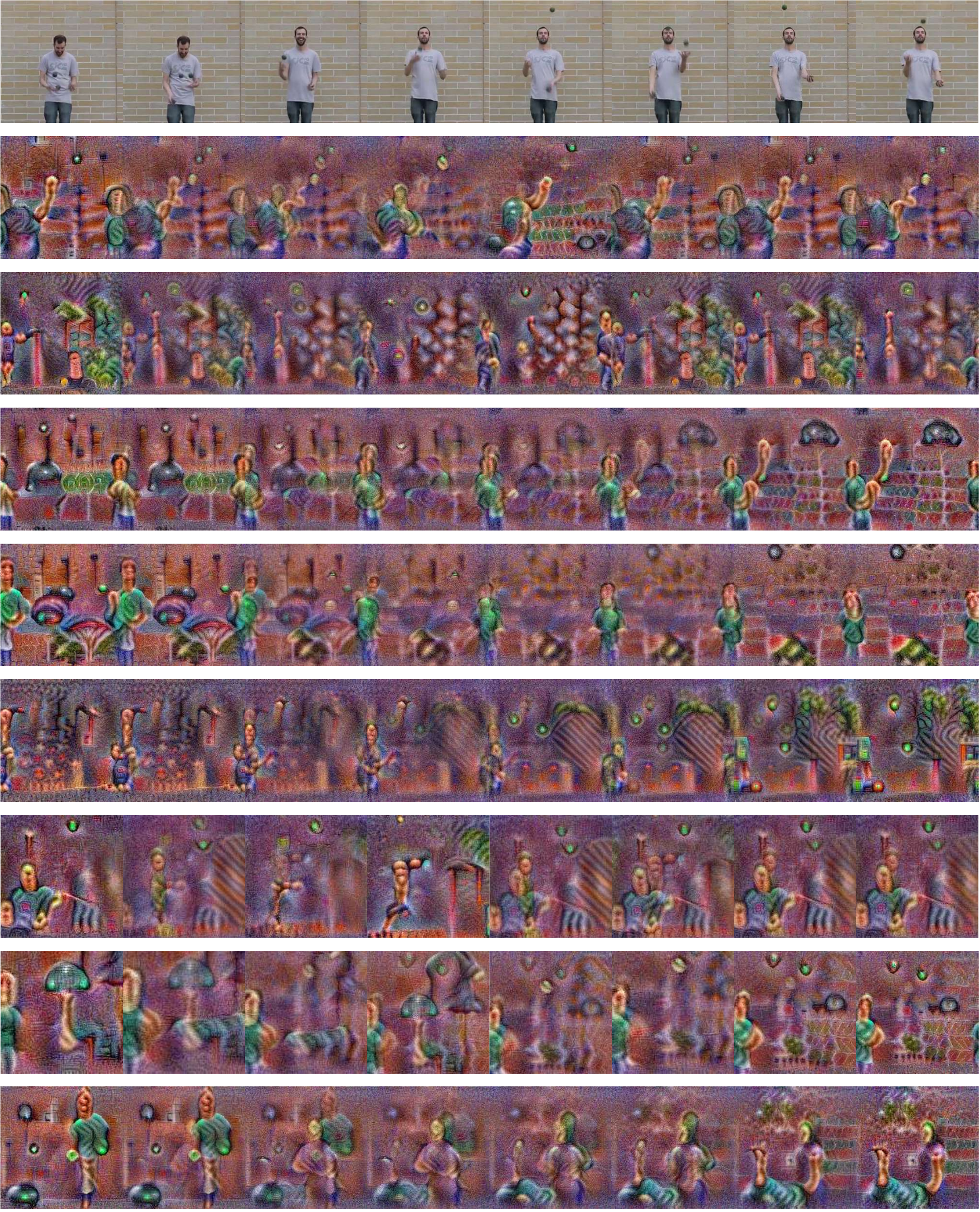}

    \put (0,94.4) {\rotatebox[origin=c]{90}{\colorbox{black}{\small{\textcolor{white}{$\;\!$stimulus video$\;\!$}}}}}

    \put (0,83.3) {\rotatebox[origin=c]{90}{\colorbox{black}{\small{\textcolor{white}{$\;\;$3D R50~\cite{hara2018can}$\quad$}}}}}

    \put (0,72.2) {\rotatebox[origin=c]{90}{\colorbox{black}{\footnotesize{\textcolor{white}{$\;$(2+1)D R50~\cite{tran2018closer}$\;$}}}}}

    \put (0,60.9) {\rotatebox[origin=c]{90}{\colorbox{black}{\small{\textcolor{white}{$\;\;$CSN R50~\cite{tran2019video}$\;$}}}}}

    \put (0,49.7) {\rotatebox[origin=c]{90}{\colorbox{black}{\small{\textcolor{white}{$\quad\!\!$X3D$_{XS}$~\cite{feichtenhofer2020x3d}$\quad\!$}}}}}

    \put (0,38.5) {\rotatebox[origin=c]{90}{\colorbox{black}{\small{\textcolor{white}{$\quad$X3D$_{S}$~\cite{feichtenhofer2020x3d}$\quad\,$}}}}}

    \put (0,27.3) {\rotatebox[origin=c]{90}{\colorbox{black}{\small{\textcolor{white}{$\quad\!$X3D$_{M}$~\cite{feichtenhofer2020x3d}$\quad$}}}}}

    \put (0,16.1) {\rotatebox[origin=c]{90}{\colorbox{black}{\small{\textcolor{white}{$\quad$X3D$_{L}$~\cite{feichtenhofer2020x3d}$\quad\,$}}}}}

    \put (0,4.9) {\rotatebox[origin=c]{90}{\colorbox{black}{\footnotesize{\textcolor{white}{TimeSformer~\cite{bertasius2021space}$\,$}}}}}
    
    \end{overpic}
    \caption{\textbf{Qualitative examples of synthesized features with LEAPS} for action label \emph{juggling balls}.}
    \label{fig:juggling_1}
\vspace{-1.1em}
\end{figure*}

\begin{figure*}[t]
\ContinuedFloat
    \centering
    \begin{overpic}[trim = {0cm, 0cm, 0cm, 0cm}, clip,width=\textwidth]
    {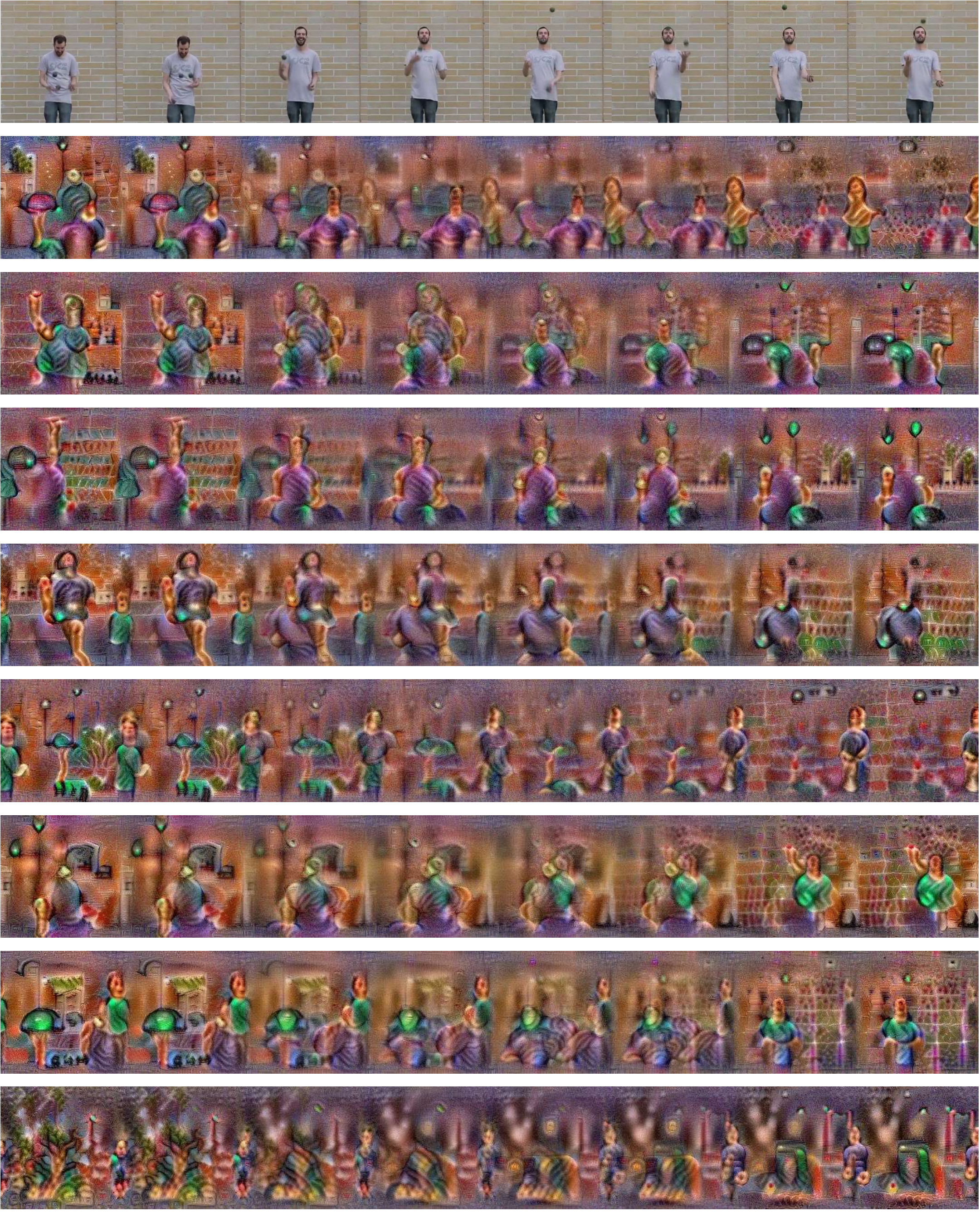}

    \put (0,94.4) {\rotatebox[origin=c]{90}{\colorbox{black}{\small{\textcolor{white}{$\;\!$stimulus video$\;\!$}}}}}

    \put (0,83.3) {\rotatebox[origin=c]{90}{\colorbox{black}{\footnotesize{\textcolor{white}{Video Swin-T~\cite{liu2022video}}}}}}

    \put (0,72.2) {\rotatebox[origin=c]{90}{\colorbox{black}{\footnotesize{\textcolor{white}{Video Swin-S~\cite{liu2022video}}}}}}

    \put (0,60.9) {\rotatebox[origin=c]{90}{\colorbox{black}{\footnotesize{\textcolor{white}{Video Swin-B~\cite{liu2022video}}}}}}

    \put (0,49.7) {\rotatebox[origin=c]{90}{\colorbox{black}{\small{\textcolor{white}{$\,$MViTv2-S~\cite{li2022mvitv2}$\;$}}}}}

    \put (0,38.5) {\rotatebox[origin=c]{90}{\colorbox{black}{\small{\textcolor{white}{$\,$MViTv2-B~\cite{li2022mvitv2}$\;$}}}}}

    \put (0,27.3) {\rotatebox[origin=c]{90}{\colorbox{black}{\small{\textcolor{white}{rev-MViT-B~\cite{mangalam2022reversible}}}}}}

    \put (0,16.2) {\rotatebox[origin=c]{90}{\colorbox{black}{\scriptsize{\textcolor{white}{UniFormerv2-B~\cite{li2022uniformerv2}$\,$}}}}}

    \put (0,5.0) {\rotatebox[origin=c]{90}{\colorbox{black}{\scriptsize{\textcolor{white}{UniFormerv2-L~\cite{li2022uniformerv2}$\,$}}}}}
    
    \end{overpic}
    \caption{\textbf{Qualitative examples of synthesized features with LEAPS} for action label \emph{juggling balls} (continued).}
    \label{fig:juggling_2}
\vspace{-1.1em}
\end{figure*}

\begin{figure*}[t]
    \centering
    \begin{overpic}[trim = {0cm, 0cm, 0cm, 0cm}, clip,width=\textwidth]
    {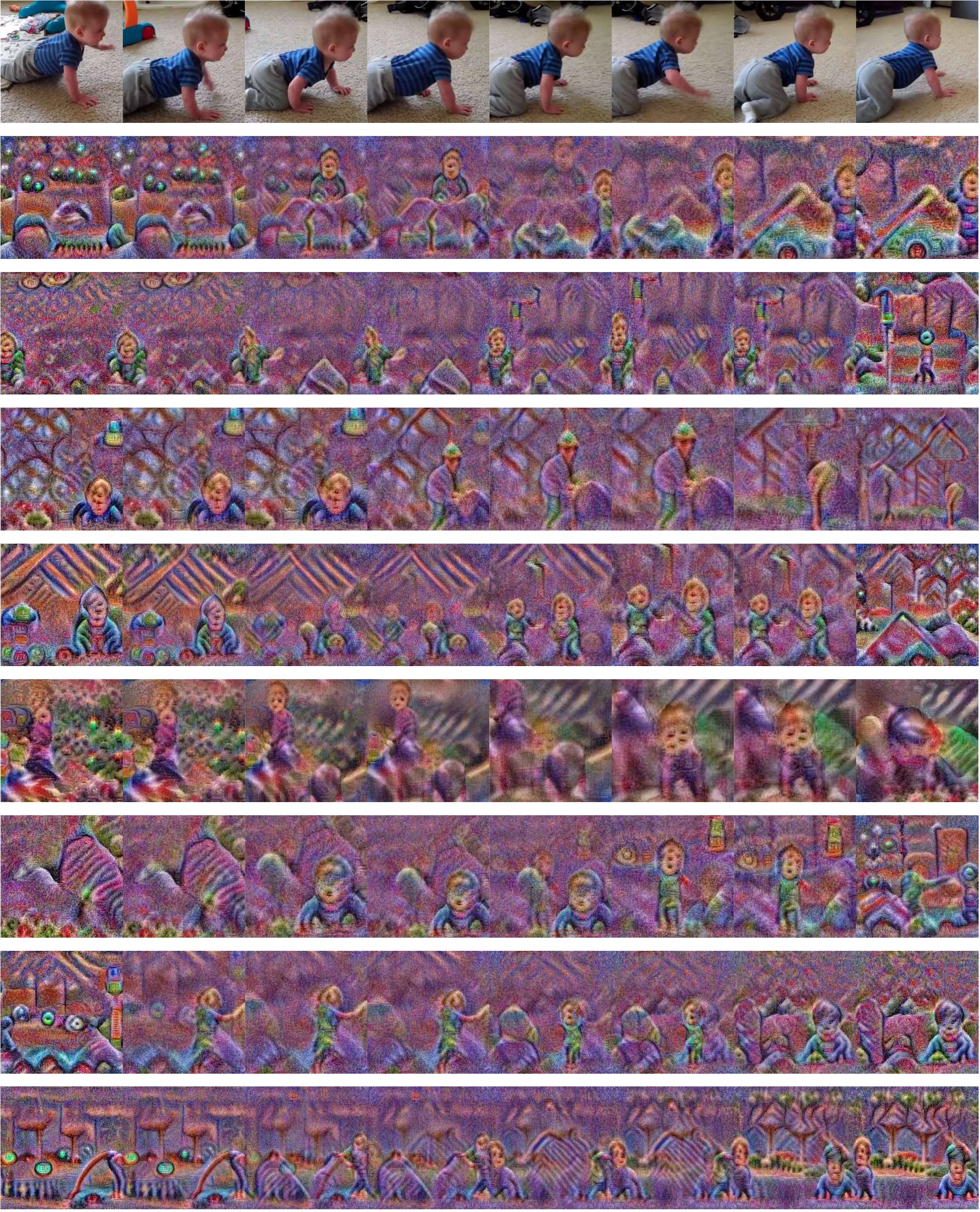}

    \put (0,94.4) {\rotatebox[origin=c]{90}{\colorbox{black}{\small{\textcolor{white}{$\;\!$stimulus video$\;\!$}}}}}

    \put (0,83.3) {\rotatebox[origin=c]{90}{\colorbox{black}{\small{\textcolor{white}{$\;\;$3D R50~\cite{hara2018can}$\quad$}}}}}

    \put (0,72.2) {\rotatebox[origin=c]{90}{\colorbox{black}{\footnotesize{\textcolor{white}{$\;$(2+1)D R50~\cite{tran2018closer}$\;$}}}}}

    \put (0,60.9) {\rotatebox[origin=c]{90}{\colorbox{black}{\small{\textcolor{white}{$\;\;$CSN R50~\cite{tran2019video}$\;$}}}}}

    \put (0,49.7) {\rotatebox[origin=c]{90}{\colorbox{black}{\small{\textcolor{white}{$\quad\!\!$X3D$_{XS}$~\cite{feichtenhofer2020x3d}$\quad\!$}}}}}

    \put (0,38.5) {\rotatebox[origin=c]{90}{\colorbox{black}{\small{\textcolor{white}{$\quad$X3D$_{S}$~\cite{feichtenhofer2020x3d}$\quad\,$}}}}}

    \put (0,27.3) {\rotatebox[origin=c]{90}{\colorbox{black}{\small{\textcolor{white}{$\quad\!$X3D$_{M}$~\cite{feichtenhofer2020x3d}$\quad$}}}}}

    \put (0,16.1) {\rotatebox[origin=c]{90}{\colorbox{black}{\small{\textcolor{white}{$\quad$X3D$_{L}$~\cite{feichtenhofer2020x3d}$\quad\,$}}}}}

    \put (0,4.9) {\rotatebox[origin=c]{90}{\colorbox{black}{\footnotesize{\textcolor{white}{TimeSformer~\cite{bertasius2021space}$\,$}}}}}
    
    \end{overpic}
    \caption{\textbf{Qualitative examples of synthesized features with LEAPS} for action label \emph{baby crawling}.}
    \label{fig:crawling}
\vspace{-1.1em}
\end{figure*}

\begin{figure*}[t]
\ContinuedFloat
    \centering
    \begin{overpic}[trim = {0cm, 0cm, 0cm, 0cm}, clip,width=\textwidth]
    {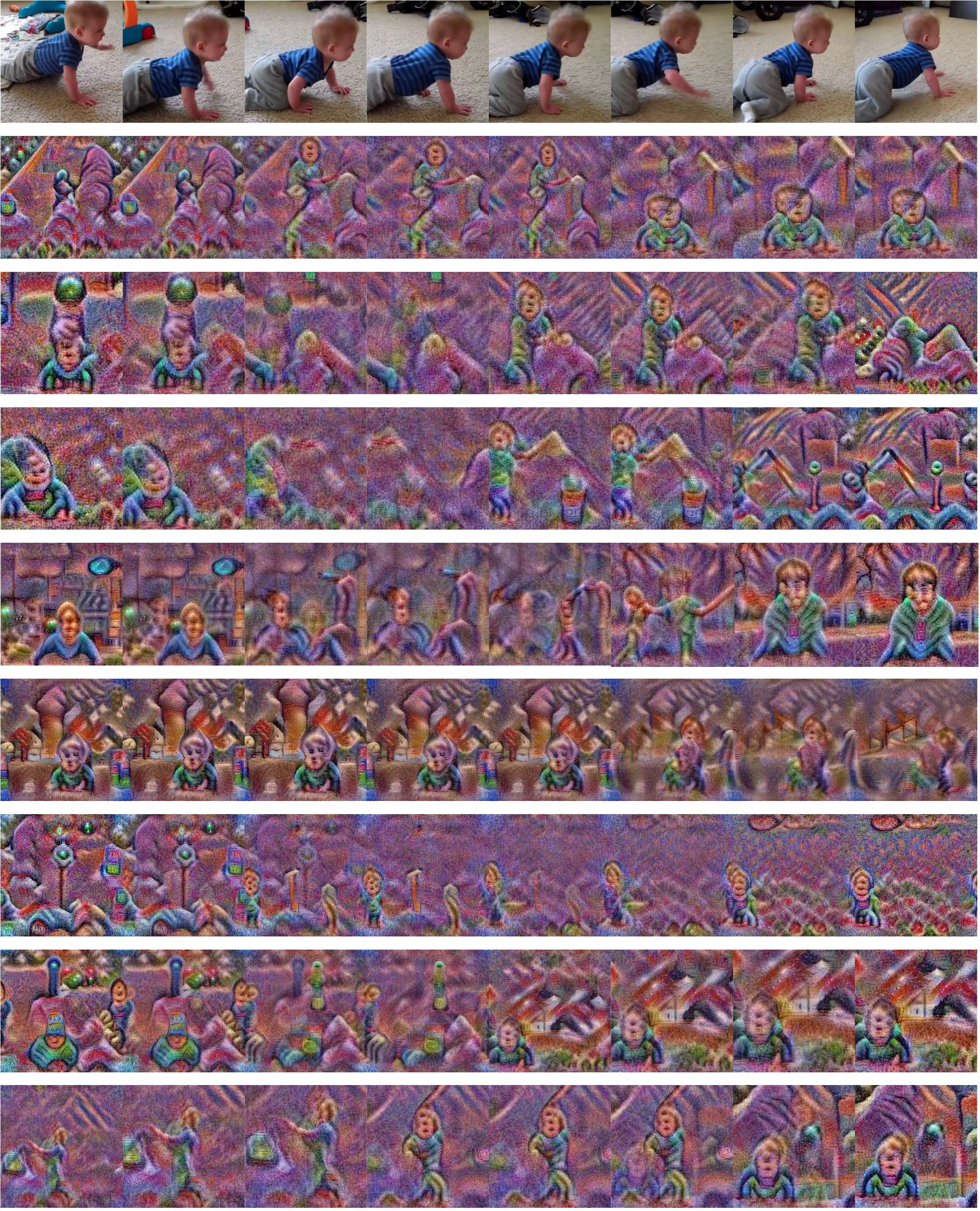}

    \put (0,94.4) {\rotatebox[origin=c]{90}{\colorbox{black}{\small{\textcolor{white}{$\;\!$stimulus video$\;\!$}}}}}

    \put (0,83.3) {\rotatebox[origin=c]{90}{\colorbox{black}{\footnotesize{\textcolor{white}{Video Swin-T~\cite{liu2022video}}}}}}

    \put (0,72.2) {\rotatebox[origin=c]{90}{\colorbox{black}{\footnotesize{\textcolor{white}{Video Swin-S~\cite{liu2022video}}}}}}

    \put (0,60.9) {\rotatebox[origin=c]{90}{\colorbox{black}{\footnotesize{\textcolor{white}{Video Swin-B~\cite{liu2022video}}}}}}

    \put (0,49.7) {\rotatebox[origin=c]{90}{\colorbox{black}{\small{\textcolor{white}{$\,$MViTv2-S~\cite{li2022mvitv2}$\;$}}}}}

    \put (0,38.5) {\rotatebox[origin=c]{90}{\colorbox{black}{\small{\textcolor{white}{$\,$MViTv2-B~\cite{li2022mvitv2}$\;$}}}}}

    \put (0,27.3) {\rotatebox[origin=c]{90}{\colorbox{black}{\small{\textcolor{white}{rev-MViT-B~\cite{mangalam2022reversible}}}}}}

    \put (0,16.2) {\rotatebox[origin=c]{90}{\colorbox{black}{\scriptsize{\textcolor{white}{UniFormerv2-B~\cite{li2022uniformerv2}$\,$}}}}}

    \put (0,5.0) {\rotatebox[origin=c]{90}{\colorbox{black}{\scriptsize{\textcolor{white}{UniFormerv2-L~\cite{li2022uniformerv2}$\,$}}}}}
    
    \end{overpic}
    \caption{\textbf{Qualitative examples of synthesized features with LEAPS} for action label \emph{baby crawling} (continued).}
    \label{fig:crawling_2}
\vspace{-1.1em}
\end{figure*}

\begin{figure*}[t]
    \centering
    \begin{overpic}[trim = {0cm, 0cm, 0cm, 0cm}, clip,width=\textwidth]
    {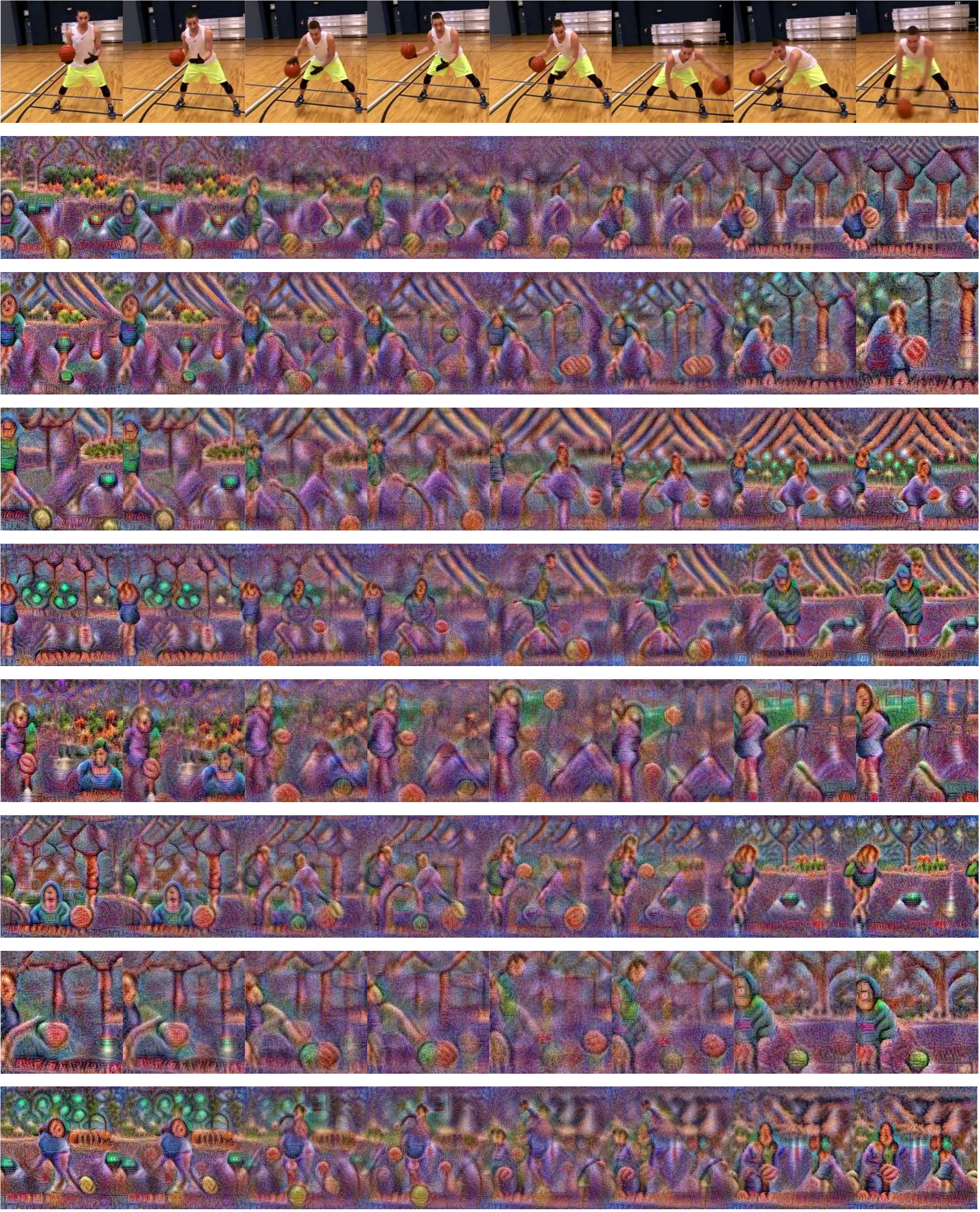}

    \put (0,94.4) {\rotatebox[origin=c]{90}{\colorbox{black}{\small{\textcolor{white}{$\;\!$stimulus video$\;\!$}}}}}

    \put (0,83.3) {\rotatebox[origin=c]{90}{\colorbox{black}{\small{\textcolor{white}{$\;\;$3D R50~\cite{hara2018can}$\quad$}}}}}

    \put (0,72.2) {\rotatebox[origin=c]{90}{\colorbox{black}{\footnotesize{\textcolor{white}{$\;$(2+1)D R50~\cite{tran2018closer}$\;$}}}}}

    \put (0,60.9) {\rotatebox[origin=c]{90}{\colorbox{black}{\small{\textcolor{white}{$\;\;$CSN R50~\cite{tran2019video}$\;$}}}}}

    \put (0,49.7) {\rotatebox[origin=c]{90}{\colorbox{black}{\small{\textcolor{white}{$\quad\!\!$X3D$_{XS}$~\cite{feichtenhofer2020x3d}$\quad\!$}}}}}

    \put (0,38.5) {\rotatebox[origin=c]{90}{\colorbox{black}{\small{\textcolor{white}{$\quad$X3D$_{S}$~\cite{feichtenhofer2020x3d}$\quad\,$}}}}}

    \put (0,27.3) {\rotatebox[origin=c]{90}{\colorbox{black}{\small{\textcolor{white}{$\quad\!$X3D$_{M}$~\cite{feichtenhofer2020x3d}$\quad$}}}}}

    \put (0,16.1) {\rotatebox[origin=c]{90}{\colorbox{black}{\small{\textcolor{white}{$\quad$X3D$_{L}$~\cite{feichtenhofer2020x3d}$\quad\,$}}}}}

    \put (0,4.9) {\rotatebox[origin=c]{90}{\colorbox{black}{\footnotesize{\textcolor{white}{TimeSformer~\cite{bertasius2021space}$\,$}}}}}
    
    \end{overpic}
    \caption{\textbf{Qualitative examples of synthesized features with LEAPS} for action label \emph{dribbling basketball}.}
    \label{fig:dribbling}
\vspace{-1.1em}
\end{figure*}

\begin{figure*}[t]
\ContinuedFloat
    \centering
    \begin{overpic}[trim = {0cm, 0cm, 0cm, 0cm}, clip,width=\textwidth]
    {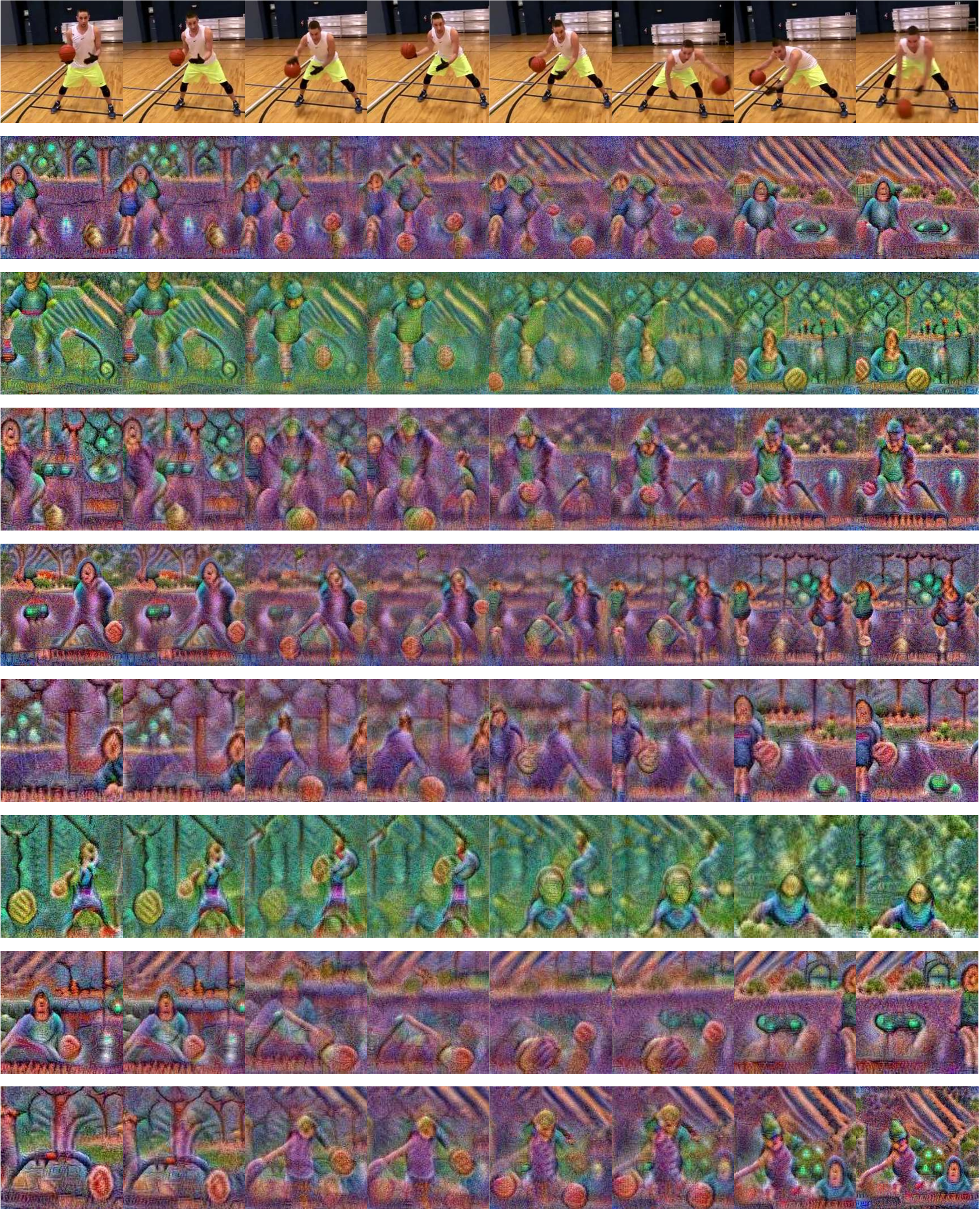}

    \put (0,94.4) {\rotatebox[origin=c]{90}{\colorbox{black}{\small{\textcolor{white}{$\;\!$stimulus video$\;\!$}}}}}

    \put (0,83.3) {\rotatebox[origin=c]{90}{\colorbox{black}{\footnotesize{\textcolor{white}{Video Swin-T~\cite{liu2022video}}}}}}

    \put (0,72.2) {\rotatebox[origin=c]{90}{\colorbox{black}{\footnotesize{\textcolor{white}{Video Swin-S~\cite{liu2022video}}}}}}

    \put (0,60.9) {\rotatebox[origin=c]{90}{\colorbox{black}{\footnotesize{\textcolor{white}{Video Swin-B~\cite{liu2022video}}}}}}

    \put (0,49.7) {\rotatebox[origin=c]{90}{\colorbox{black}{\small{\textcolor{white}{$\,$MViTv2-S~\cite{li2022mvitv2}$\;$}}}}}

    \put (0,38.5) {\rotatebox[origin=c]{90}{\colorbox{black}{\small{\textcolor{white}{$\,$MViTv2-B~\cite{li2022mvitv2}$\;$}}}}}

    \put (0,27.3) {\rotatebox[origin=c]{90}{\colorbox{black}{\small{\textcolor{white}{rev-MViT-B~\cite{mangalam2022reversible}}}}}}

    \put (0,16.2) {\rotatebox[origin=c]{90}{\colorbox{black}{\scriptsize{\textcolor{white}{UniFormerv2-B~\cite{li2022uniformerv2}$\,$}}}}}

    \put (0,5.0) {\rotatebox[origin=c]{90}{\colorbox{black}{\scriptsize{\textcolor{white}{UniFormerv2-L~\cite{li2022uniformerv2}$\,$}}}}}
    
    \end{overpic}
    \caption{\textbf{Qualitative examples of synthesized features with LEAPS} for action label \emph{dribbling basketball} (continued).}
    \label{fig:dribbling_2}
\vspace{-1.1em}
\end{figure*}

\begin{figure*}[t]
    \centering
    \begin{overpic}[trim = {0cm, 0cm, 0cm, 0cm}, clip,width=\textwidth]
    {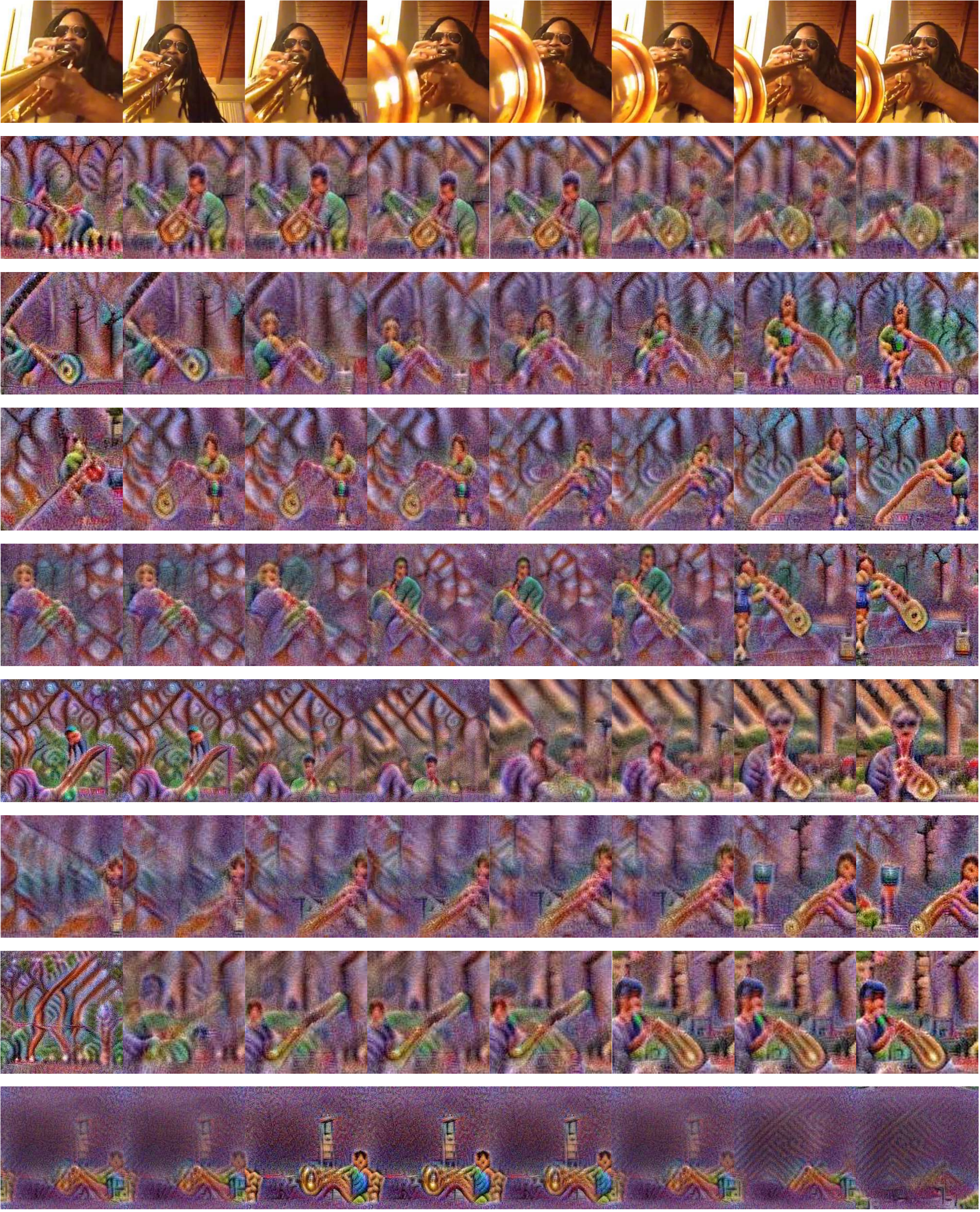}

    \put (0,94.4) {\rotatebox[origin=c]{90}{\colorbox{black}{\small{\textcolor{white}{$\;\!$stimulus video$\;\!$}}}}}

    \put (0,83.3) {\rotatebox[origin=c]{90}{\colorbox{black}{\small{\textcolor{white}{$\;\;$3D R50~\cite{hara2018can}$\quad$}}}}}

    \put (0,72.2) {\rotatebox[origin=c]{90}{\colorbox{black}{\footnotesize{\textcolor{white}{$\;$(2+1)D R50~\cite{tran2018closer}$\;$}}}}}

    \put (0,60.9) {\rotatebox[origin=c]{90}{\colorbox{black}{\small{\textcolor{white}{$\;\;$CSN R50~\cite{tran2019video}$\;$}}}}}

    \put (0,49.7) {\rotatebox[origin=c]{90}{\colorbox{black}{\small{\textcolor{white}{$\quad\!\!$X3D$_{XS}$~\cite{feichtenhofer2020x3d}$\quad\!$}}}}}

    \put (0,38.5) {\rotatebox[origin=c]{90}{\colorbox{black}{\small{\textcolor{white}{$\quad$X3D$_{S}$~\cite{feichtenhofer2020x3d}$\quad\,$}}}}}

    \put (0,27.3) {\rotatebox[origin=c]{90}{\colorbox{black}{\small{\textcolor{white}{$\quad\!$X3D$_{M}$~\cite{feichtenhofer2020x3d}$\quad$}}}}}

    \put (0,16.1) {\rotatebox[origin=c]{90}{\colorbox{black}{\small{\textcolor{white}{$\quad$X3D$_{L}$~\cite{feichtenhofer2020x3d}$\quad\,$}}}}}

    \put (0,4.9) {\rotatebox[origin=c]{90}{\colorbox{black}{\footnotesize{\textcolor{white}{TimeSformer~\cite{bertasius2021space}$\,$}}}}}
    
    \end{overpic}
    \caption{\textbf{Qualitative examples of synthesized features with LEAPS} for action label \emph{playing trumpet}.}
    \label{fig:trumpet}
\vspace{-1.1em}
\end{figure*}

\begin{figure*}[t]
\ContinuedFloat
    \centering
    \begin{overpic}[trim = {0cm, 0cm, 0cm, 0cm}, clip,width=\textwidth]
    {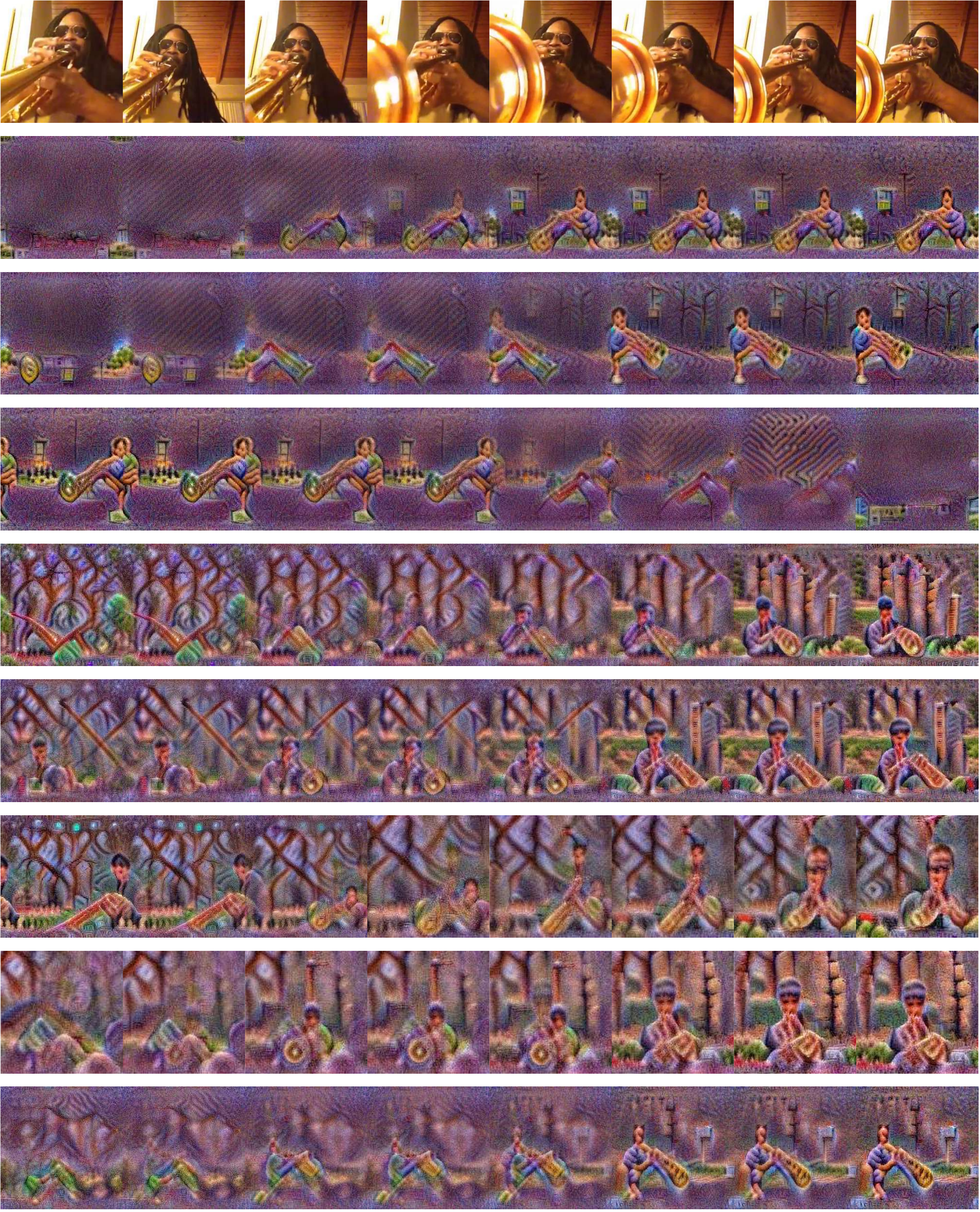}

    \put (0,94.4) {\rotatebox[origin=c]{90}{\colorbox{black}{\small{\textcolor{white}{$\;\!$stimulus video$\;\!$}}}}}

    \put (0,83.3) {\rotatebox[origin=c]{90}{\colorbox{black}{\footnotesize{\textcolor{white}{Video Swin-T~\cite{liu2022video}}}}}}

    \put (0,72.2) {\rotatebox[origin=c]{90}{\colorbox{black}{\footnotesize{\textcolor{white}{Video Swin-S~\cite{liu2022video}}}}}}

    \put (0,60.9) {\rotatebox[origin=c]{90}{\colorbox{black}{\footnotesize{\textcolor{white}{Video Swin-B~\cite{liu2022video}}}}}}

    \put (0,49.7) {\rotatebox[origin=c]{90}{\colorbox{black}{\small{\textcolor{white}{$\,$MViTv2-S~\cite{li2022mvitv2}$\;$}}}}}

    \put (0,38.5) {\rotatebox[origin=c]{90}{\colorbox{black}{\small{\textcolor{white}{$\,$MViTv2-B~\cite{li2022mvitv2}$\;$}}}}}

    \put (0,27.3) {\rotatebox[origin=c]{90}{\colorbox{black}{\small{\textcolor{white}{rev-MViT-B~\cite{mangalam2022reversible}}}}}}

    \put (0,16.2) {\rotatebox[origin=c]{90}{\colorbox{black}{\scriptsize{\textcolor{white}{UniFormerv2-B~\cite{li2022uniformerv2}$\,$}}}}}

    \put (0,5.0) {\rotatebox[origin=c]{90}{\colorbox{black}{\scriptsize{\textcolor{white}{UniFormerv2-L~\cite{li2022uniformerv2}$\,$}}}}}
    
    \end{overpic}
    \caption{\textbf{Qualitative examples of synthesized features with LEAPS} for action label \emph{playing trumpet} (continued).}
    \label{fig:trumpet_2}
\vspace{-1.1em}
\end{figure*}

\section{Hyperparameter settings}

As described in Section~\textcolor{red}{4.1}, we discover the optimal $\lambda$ and $r$ hyperparameters for each model through grid search. To limit the search space and computational overhead of hyperparameter tuning, we define $\lambda_1 \! \in \! \{ 0.5, 0.625, 0.75, 0.875, 1.0 \}$,
$\lambda_L \! \in \! \{ 0.1, 0.2, 0.3, 0.4, 0.5 \}$,
$r \! \in \! \{ 1e^{-3}, 2.5e^{-3}, 5e^{-3}, 7.5^{e-3}, 1e^{-2} \}$, where $\lambda_1$ is the priming weight for the first layer of the network, $\lambda_L$ is the priming weight for the final layer of the network. Based on $\lambda_1$ and $\lambda_L$, we use a linear (decreasing) function for the remaining $\lambda \in \{2,...,L-1\}$ layer priming weights. \Cref{tab:hyperparams} provides a full list of the hyperparameters discovered and used for inverting each model. We note that the loss shows to increase in larger models due to the number of layers used for priming.

\begin{figure}[t]
    \centering
    \begin{overpic}[width=\linewidth]{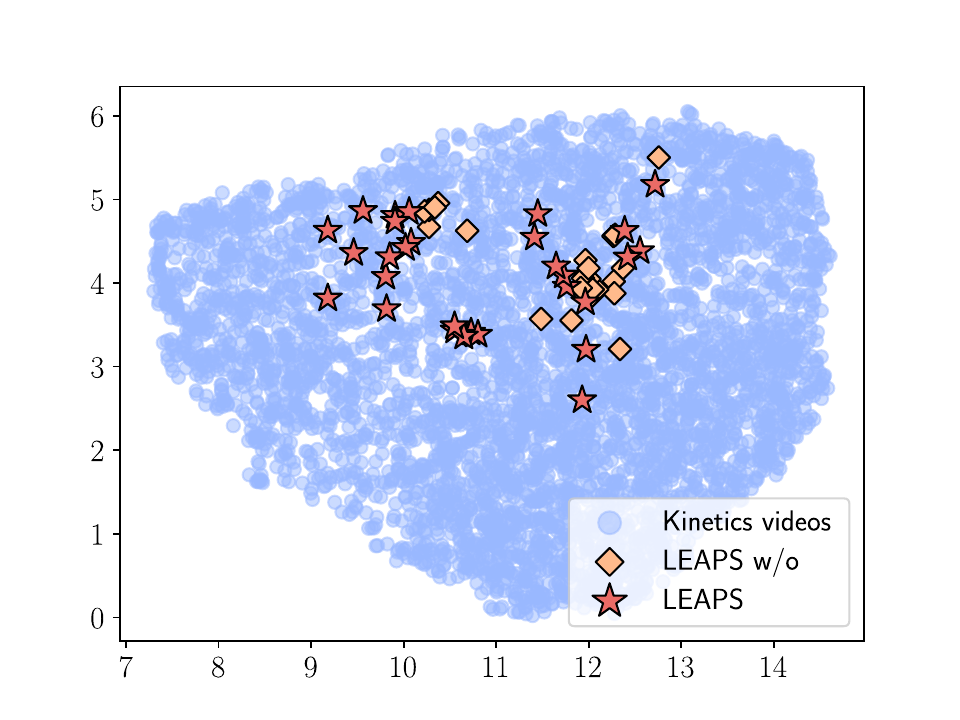}

    \put (30,0) {\small{First principal component}}
    \put (3.5,35.0) {\rotatebox[origin=c]{90}{\small{Second principal component}}}
    \put (82.5,16) {\scriptsize{$\underset{feat}{\mathcal{R}}$}}

    \end{overpic}
    \caption{\textbf{Projection of X3D}$_{\textbf{M}}$\textbf{'s final encoder layer embeddings} onto two principal components for Kinetic's class \emph{tai chi}. Embeddings of videos from Kinetics are in \textcolor[HTML]{4B67A6}{blue}, from LEAPS w/o $\underset{feat}{\mathcal{R}}$ in \textcolor[HTML]{FFBA8C}{orange}, and from LEAPS in \textcolor[HTML]{DB5C57}{red}.}
    \label{fig:umap}
\end{figure}

\section{Embedding space visualizations}

LEAPS aims to synthesize visually coherent representations of inverted models. To better understand the relationship between the inverted model's features and real videos from the Kinetics-400 train set, we provide UMAP~\cite{mcinnes2018umap} visualizations of their feature embeddings for action \emph{tai chi}. We use the spatiotemporally averaged feature vectors from the final convolution block in X3D$_{\text{M}}$ (\texttt{s5.pathway0\_res6.branch2.c}).

As illustrated from the results in~\Cref{fig:umap}, inverted model embeddings are within the distribution of embeddings from Kinetics videos. While this is true for both embeddings from LEAPS synthesized videos as well as LEAPS synthesized videos without feature diversity regularization, LEAPS videos show a greater level of variation without being as closely concentrated as the embeddings of LEAPS w/o $\underset{feat}{\mathcal{R}}$.

\begin{table}[t]
\centering
\resizebox{\linewidth}{!}{%
\begin{tabular}{ l| c c | l l }
\hline
Priming &
\multicolumn{2}{c|}{top-1 (\%)} &
\multicolumn{2}{c}{Inception Score (IS)} \\
layers (\%) & model & ver. & model & verifier \tstrut \\
\hline
\rowcolor{LightGrey} \multicolumn{5}{l}{\textit{3D R50}} \tstrut \\
20 & 19.0 & 4.1 & 1.3 $\pm$ 0.2 & 1.1 $\pm$ 0.1 \\
40 & 23.4 & 9.5 & 1.8 $\pm$ 0.4 & 1.4 $\pm$ 0.4 \\
60 & 41.8 & 23.4 & 2.5 $\pm$ 0.6 & 1.6 $\pm$ 0.5 \\
80 & 69.3 & 54.6 & 4.2 $\pm$ 1.3 & 2.0 $\pm$ 0.4 \\
100 (LEAPS) & \textbf{86.7} & \textbf{68.5} & \textbf{9.0} $\pm$ \textbf{1.0} & \textbf{5.7} $\pm$ \textbf{0.7} \\
\hline
\rowcolor{LightGrey} \multicolumn{5}{l}{\textit{X3D$_{\text{M}}$}} \tstrut \\
20 & 15.8 & 3.9 & 1.1 $\pm$ 0.1 & 1.0 \\
40 & 18.3 & 5.4 & 1.4 $\pm$ 0.4 & 1.0  \\
60 & 32.6 & 18.7 & 2.1 $\pm$ 0.8 & 1.2 $\pm$ 0.2 \\
80 & 55.0 & 37.2 & 3.8 $\pm$ 0.7 & 2.1 $\pm$ 0.6 \\
100 (LEAPS) & \textbf{90.3} & \textbf{82.5} & \textbf{11.4} $\pm$ \textbf{0.9} & \textbf{8.0} $\pm$ \textbf{1.4} \\
\end{tabular}
}
\caption{\textbf{Ablation on the percentage of model's layers used for priming}. The best results per metric are in \textbf{bold}.}
\label{tab:priming_layers}
\vspace{-1.1em}
\end{table}

\section{Priming layers}

We further ablate over the number of layers used by the priming loss $\underset{prim}{\mathcal{L}}$.  We select embeddings from the first 20\%, 40\%, 60\%, and 80\% of the total network layers for our priming loss. Given our proposed LEAPS uses embeddings from all network layers; i.e. $\mathbf{\Lambda}=\{1,...,L\}$, each setting in turn uses $\mathbf{\Lambda}_{20}=\{1,..., \lfloor \frac{L}{5} \rfloor \}$, $\mathbf{\Lambda}_{40}=\{1,...,\frac{2L}{5} \rfloor \}$,
$\mathbf{\Lambda}_{60}=\{1,...,\lfloor\frac{3L}{5} \rfloor \}$, and
$\mathbf{\Lambda}_{80}=\{1,...,\lfloor\frac{4L}{5} \rfloor \}$, where $\lfloor \cdot \rfloor$ denotes the floor function. As shown in \Cref{tab:priming_layers}, for both 3D R50 and X3D$_{\text{M}}$, priming layer reductions also correspond to large decreases in top-1 accuracies and inception scores. The degradation in accuracy and IS is observed for both the inverted models as well as the verifier. 

\begin{figure}
     \centering
     \includegraphics[width=\linewidth]{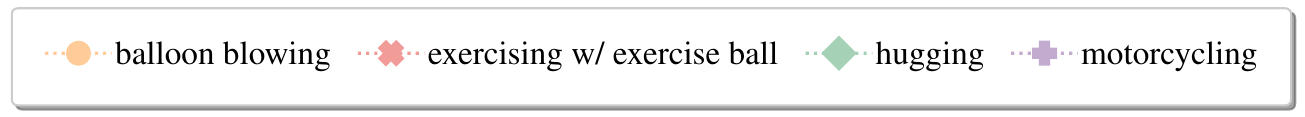}
     \begin{subfigure}[b]{0.49\linewidth}
         \centering
         \includegraphics[width=\textwidth, trim={.8cm .8cm 0cm 0cm},clip]{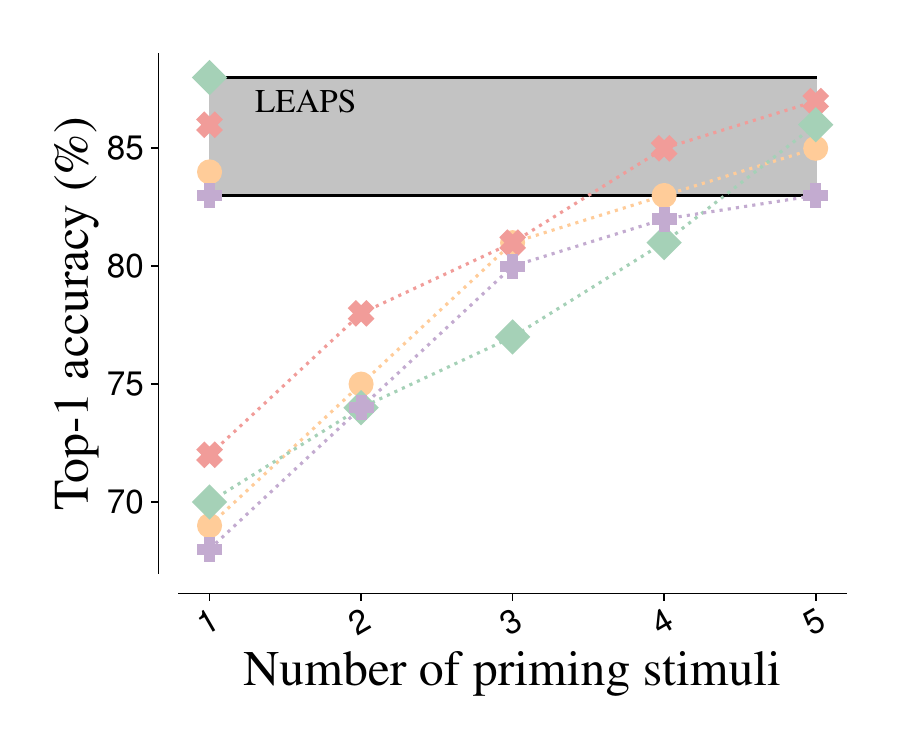}
         \caption{3D R50}
         \label{fig:y equals x}
     \end{subfigure}
     \hfill
     \begin{subfigure}[b]{0.49\linewidth}
         \centering
         \includegraphics[width=\textwidth, trim={.8cm .8cm 0cm 0cm},clip]{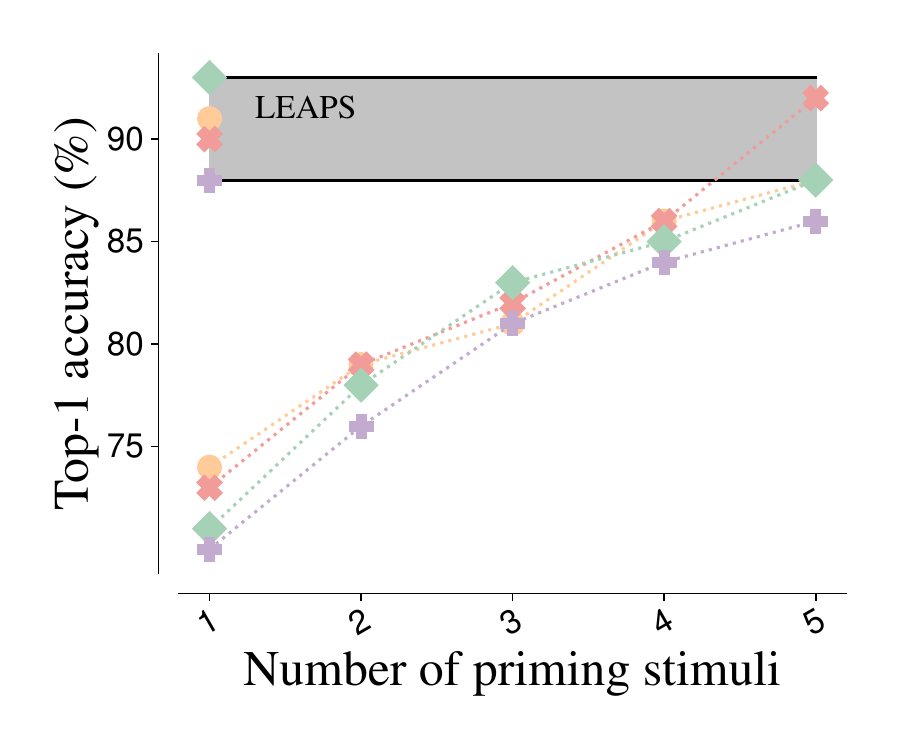}
         \caption{X3D$_{\text{M}}$}
         \label{fig:three sin x}
     \end{subfigure}
        \caption{\textbf{Top-1 inverted model accuracy (\%) with priming} over stimuli videos. The area between the lower and upper class-accuracy bounds achieved by videos from LEAPS is shown in \textcolor[HTML]{c3c3c3}{gray}.}
        \label{fig:multi_stimuli_acc}
\end{figure}

\section{Multi-stimuli priming}
\label{sec:multi_simuli}
Our proposed video model inversion method is based on the approximation of embeddings that are relevant to specific actions. LEAPS uses the embeddings from a single priming example as stimulus. As an alternative, one may use additional stimuli videos to recall the learned preconscious of models associated with a class. We show in \Cref{fig:multi_stimuli_acc} the top-1 accuracies achieved by 3D R50 and X3D$_{\text{M}}$ when priming is performed with multiple stimuli instead of using LEAPS regularizers. As observed, the use of temporal coherence and feature diversity regularizers terms can perform favorably over internal representations from a small number of multiple stimuli. However, increasing the number of stimuli used show comparable performance to that achieved by LEAPS, thus advocating for an alternative to regularizers when access to more data is available.    

\begin{figure*}
    \centering
    \includegraphics[width=\textwidth]{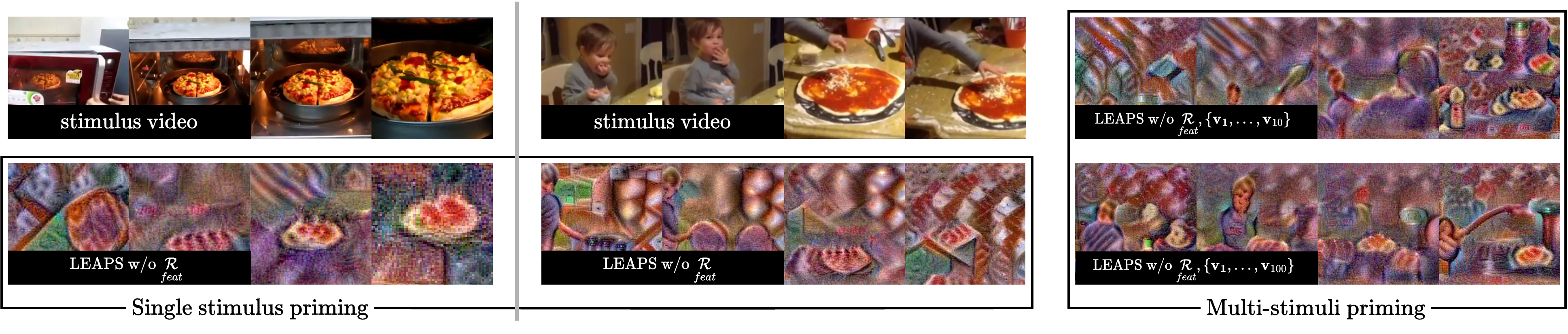}
    \caption{\textbf{Single and multi stimuli priming without} $\underset{feat}{\mathcal{R}}$ \textbf{for class} \emph{making pizza}. The leftmost and center columns use a single but different stimulus video for model inversion. The right column uses the mean embeddings over 10 (top) and 100 (bottom) stimuli videos of the corresponding class. MViTv2-B features are inverted without a verifier network.}
    \vspace{-4pt}
\label{fig:leaps_reg_ablations}
\end{figure*}

\section{Additional Discussions}

\noindent
\textbf{Limitations}. LEAPS is a general model-independent method for visualizing learned concepts of video models. We have demonstrated its effectiveness in inverting multiple architectures. As the synthesized visual features are not influenced by training data, with only a single stimulus video used to prime the network, we include a feature diversity regularizer. The regularizer uses the batch norm statistics as in~\cite{yin2020dreaming}, to approximate realistic features given a verifier network. The verifier is limited to architectures with batch norm layers and restricts the use of attention-based models.

We consider two approaches to mitigate this. The first approach is to remove the diversity regularized altogether. This evidently results in accuracy and IS decrease as shown in Table~\textcolor{red}{3} with \textbf{LEAPS} $\underset{prim}{\mathcal{L}} + \underset{coh}{\mathcal{R}}$ and \textbf{LEAPS (full)}. Qualitative examples are shown in the left and middle columns of \Cref{fig:leaps_reg_ablations}. The second approach is the use of multi-stimuli priming, which shows promise as an alternative in settings where additional data is available, as discussed in Section~\ref{sec:multi_simuli}. We also provide examples of the effect of multi-stimuli priming at the rightmost column of \Cref{fig:leaps_reg_ablations}.

\noindent
\textbf{Applicability to other tasks}. Our focus has been on the inversion of video models and the visualization of their embeddings. However, the method can be further extended to subsequent downstream tasks in the video domain including knowledge transfer~\cite{yin2020dreaming}, domain adaptation~\cite{liu2021source}, counterfactual explanations~\cite{thiagarajan2021designing}
, and inversion attacks~\cite{hatamizadeh2022gradvit}. Such tasks have received little attention for video inputs and thus we believe that LEAPS can enable subsequent research efforts.

\end{document}